\documentclass[journal,twoside,web]{ieeecolor}
\usepackage{generic}
\usepackage{amsmath,amssymb,amsfonts}

\usepackage{textcomp}
\usepackage{graphicx, url}
\usepackage{caption}
\usepackage{cite}
\usepackage[ruled, vlined]{algorithm2e}
\usepackage{subfigure}
\usepackage{subcaption}
\usepackage{tabularx,booktabs}

\usepackage{ifpdf}
 \ifpdf
 \else
 \fi

\usepackage{cite}

\usepackage{amsthm}

\newtheoremstyle{mystyle}
  {}               % Space above
  {}               % Space below
  {\itshape}       % Body font
  {}               % Indent amount
  {\bfseries}      % Theorem head font
  {: }             % Punctuation after theorem head
  {0pt}               % Space after theorem head
  {\thmname{#1}\thmnumber{ #2}\thmnote{ (#3)}} % Theorem head spec

\theoremstyle{mystyle}
\newtheorem{definition}{Definition}
\newtheorem{theorem}{Theorem}
\newtheorem{remark}{Remark}

\usepackage{url}

\makeatletter
\let\NAT@parse\undefined
\makeatother
\usepackage{hyperref}  %hyperref still needs to be put at the end!
\usepackage{cleveref}

\def\BibTeX{{\rm B\kern-.05em{\sc i\kern-.025em b}\kern-.08em
    T\kern-.1667em\lower.7ex\hbox{E}\kern-.125emX}}
\usepackage[protrusion=true,expansion=true]{microtype}

\usepackage{balance}
\everymath{\footnotesize} % Reduce font size of inline equations
\everydisplay{\small} % Reduce font size of display equations
\usepackage{nicefrac}
\allowdisplaybreaks

\begin{document}

\title{Game-Theoretic Safe Multi-Agent Motion Planning with Reachability Analysis for Dynamic and Uncertain Environments (Extended Version)}

\author{Wenbin Mai, 
Minghui Liwang, \IEEEmembership{Senior Member, IEEE}, Xinlei Yi, \IEEEmembership{Senior Member, IEEE}, 
Xiaoyu Xia, \IEEEmembership{Senior Member, IEEE}, 
Seyyedali Hosseinalipour, \IEEEmembership{Senior Member, IEEE}, 
Xianbin Wang, \IEEEmembership{Fellow, IEEE}
\thanks{Accepted for publication in IEEE Transactions on Industrial Informatics. © 2025 IEEE. DOI: 10.1109/TII.2025.3627632}
\thanks{© 2025 IEEE.  Personal use of this material is permitted.  Permission from IEEE must be obtained for all other uses, in any current or future media, including reprinting/republishing this material for advertising or promotional purposes, creating new collective works, for resale or redistribution to servers or lists, or reuse of any copyrighted component of this work in other works. }
\thanks{This work was supported in part by the National Natural Science Foundation of China under Grant nos. 62271424, 62503365, 62088101, 72171172, 92367101; Shanghai Pujiang Programme under Grant no. 24PJD117; Shanghai Municipal Science and Technology Major Project under Grant no. 2021SHZDZX0100; the U.S. National Science Foundation (NSF) under Grant no. ECCS-2512911; the Chinese Academy of Engineering, Strategic Research and Consulting Program under Grant no. 2023-XZ-65; Chinese Academy of Engineering, Strategic Research and Consulting Projects, under Grant 2025-DFZD-27; Shanghai Explorer Plan project under Grant 24TS1400900. (Corresponding
author: Minghui Liwang.) }
\thanks{Wenbin Mai is with the Department of Electrical and Computer Engineering, National University of Singapore, Singapore 117583 (e-mail: wenbin.mai@u.nus.edu).}
\thanks{Minghui Liwang and Xinlei Yi are with the Department of Control Science and Engineering, Shanghai Institute of Intelligent Science and Technology, the National Key Laboratory of Autonomous Intelligent Unmanned Systems, and also with Frontiers Science Center for Intelligent Autonomous Systems, Ministry of Education, Tongji University, Shanghai 200092, China (e-mail: minghuiliwang@tongji.edu.cn; xinleiyi@tongji.edu.cn).}
\thanks{Xiaoyu Xia is with the School of Computing Technologies, RMIT University, Melbourne, Victoria, Australia (e-mail: xiaoyu.xia@rmit.edu.au). }
\thanks{Seyyedali Hosseinalipour is with the Department of Electrical Engineering, University at Buffalo-SUNY, NY, USA (e-mail: alipour@buffalo.edu). }
\thanks{Xianbin Wang is with the Department of Electrical and Computer Engineering, Western University, Ontario, Canada (e-mail: xianbin.wang@uwo.ca). }
}

\maketitle

\begin{abstract}
Ensuring safe, robust, and scalable motion planning for multi-agent systems in dynamic and uncertain environments is a persistent challenge, driven by complex inter-agent interactions, stochastic disturbances, and model uncertainties. To overcome these challenges, particularly the computational complexity of coupled decision-making and the need for proactive safety guarantees, we propose a Reachability-Enhanced Dynamic Potential Game (RE-DPG) framework, which integrates game-theoretic coordination into reachability analysis. This approach formulates multi-agent coordination as a dynamic potential game, where the Nash equilibrium (NE) defines optimal control strategies across agents. To enable scalability and decentralized execution, we develop a Neighborhood-Dominated iterative Best Response (ND-iBR) scheme, built upon an iterated $\varepsilon$-BR (i$\varepsilon$-BR) process that guarantees finite-step convergence to an $\varepsilon$-NE. This allows agents to compute strategies based on local interactions while ensuring theoretical convergence guarantees. Furthermore, to ensure safety under uncertainty, we integrate a Multi-Agent Forward Reachable Set (MA-FRS) mechanism into the cost function, explicitly modeling uncertainty propagation and enforcing collision avoidance constraints. Through both simulations and real-world experiments in 2D and 3D environments, we validate the effectiveness of RE-DPG across diverse operational scenarios.
% Ensuring safe, robust, and scalable motion planning for multi-agent systems in dynamic and uncertain environments is a persistent challenge, driven by complex inter-agent interactions, stochastic disturbances, and model uncertainties. To overcome these challenges, particularly the computational complexity of coupled decision-making and the need for proactive safety guarantees, we propose a Reachability-Enhanced Dynamic Potential Game (RE-DPG) framework, which integrates game-theoretic coordination into reachability analysis. This approach formulates multi-agent coordination as a dynamic potential game, where the Nash equilibrium (NE) defines optimal control strategies across agents. To enable scalability and decentralized execution, we develop a Neighborhood-Dominated iterative Best Response (ND-iBR) scheme, allowing agents to compute strategies based on local interactions. Furthermore, to ensure safety under uncertainty, we integrate a Multi-Agent Forward Reachable Set (MA-FRS) mechanism into the cost structure, explicitly modeling uncertainty propagation and enforcing collision avoidance constraints. Through both simulations and real-world experiments in 2D and 3D environments, we validate the effectiveness of RE-DPG across diverse operational scenarios.
\end{abstract}


% \begin{IEEEkeywords}
% Multi-agent systems, motion planning, dynamic potential game, reachability analysis, collision avoidance, distributed optimization.
% \end{IEEEkeywords}

\section{Introduction}
\IEEEPARstart{R}{ecent} advancements in robotics, artificial intelligence, and control methods have led to the widespread use of multi-agent autonomous systems in warehouse logistics \cite{turhanlar2024autonomous}, autonomous driving \cite{wang2025coordination}, and drone coordination \cite{chen2023noncooperative, niu2024integrated}. These systems operate under a shared paradigm where agents pursue goals while adapting to their environment and interacting with each other. An illustration of this interplay can be found in automated warehouse systems, where heterogeneous fleets of autonomous mobile robots must simultaneously navigate without collisions and manage dynamic tasks (e.g., prioritizing urgent orders) within constrained operational spaces.

% Recent innovations in robotics, artificial intelligence, and advanced control methodologies have catalyzed the widespread implementation of multi-agent autonomous systems across diverse domains such as intelligent warehouse logistics \cite{turhanlar2024autonomous}, autonomous driving \cite{yan2023multi, fang2024cooperative, wang2025coordination}, and distributed multi-unmanned aerial vehicle (UAV) coordination frameworks \cite{niu2024integrated, chen2023noncooperative, ruan2023hawk}. 
% Central to these applications is a shared operational paradigm in which individual agents not only pursue goal-oriented tasks but also continuously perceive their environments and adapt to interactions with neighboring agents. This dual focus on autonomy and collaboration is governed by both predefined task objectives and emergent collective behaviors arising from inter-agent communication and spatio-temporal dynamics. A clear illustration of this interplay can be found in automated warehouse systems, where heterogeneous fleets of autonomous mobile robots must simultaneously navigate collision-free trajectories and manage dynamic task allocation (e.g., prioritizing urgent orders) within constrained operational spaces. Such scenarios exemplify the inherent complexity of coordinating multiple agents in dynamic environments, highlighting the critical need for robust and adaptive motion planning strategies.

Nevertheless, real-world environments are inherently dynamic and stochastic, where external disturbances or model inaccuracies pose significant challenges to safe motion planning. Such uncertainties can lead to safety-critical situations, especially in densely populated settings where small trajectory deviations can cause cascading collisions. To ensure reliable operation, motion planning frameworks must address three key requirements: \textit{(i)} explicit modeling of inter-agent interactions, \textit{(ii)} scalability as the number of agents increases, and \textit{(iii)} ensuring safety under environmental uncertainties. Meeting these requirements presents several interconnected challenges, three instances of which are of our interest in this work:

\noindent$\bullet$ \textit{Challenge 1: Interaction Modeling.}  
In multi-agent systems, behavior of each agent depends on its own goal and the evolving influence of neighboring agents. Capturing these dynamic couplings leads to significant computational complexity, and often renders the motion planning problem  NP-hard~\cite{ARDIZZONI2024111593}.

\noindent$\bullet$ \textit{Challenge 2: Scalability under Structural Coupling.}  
Decomposing multi-agent planning into smaller subproblems is essential for scalability. However, agents share a common environment and are constrained by inter-agent safety requirements, which introduce structural coupling that complicates decentralized coordination as the system scales.

\noindent$\bullet$ \textit{Challenge 3: Uncertainty-Aware Safety Assurance.}  
In uncertain environments, deviations from nominal trajectories can lead to unsafe proximity or collisions. Ensuring robust safety thus requires modeling the propagation of disturbances and defining safety margins that remain effective under model errors and environmental variations.

Addressing the above challenges forms the core motivation of our work. To this end, we propose a multi-agent motion planning methodology named \textit{Reachability-Enhanced Dynamic Potential Game (RE-DPG)}, which integrates reachability analysis into dynamic potential game theory. This approach is designed to enable interaction-aware, safety assuring, and scalable motion planning in dynamic and uncertain environments.
\subsection{Relevant Investigations}
\subsubsection{Game-based Multi-Agent Motion Planning}
Game-theoretic frameworks provide a natural foundation for modeling strategic interactions among multiple agents \cite{hu2024time}. A central goal in these formulations is identifying a \textit{Nash equilibrium (NE)}, a stable solution where no agent can unilaterally improve its outcome by altering its strategy. 
Among the diverse classes of games, \textit{dynamic potential games (DPGs)} have emerged as particularly advantageous due to their mathematical properties, which guarantee the existence of computationally tractable NE solutions \cite{zazo2016dynamic, kavuncu2021potential, sun2024distributed}. These games offer efficient NE computation by aligning individual agent objectives with a \textit{global potential function}, thereby transforming the multi-agent coordination problems into structured optimization formulations.

In this domain, several studies have focused on centralized solvers for computing NE in DPGs. For example, \textit{Bhatt et al.} formulated a constrained optimal control problem, where solving it yields the NE for DPGs \cite{bhatt2023efficient, bhatt2025strategic}. However, centralized approaches suffer from scalability issues, as their computational demands grow with the number of agents. Additionally, these methods rely on idealized assumptions, such as \textit{perfect rationality} (i.e., agents are presumed to make fully optimal decisions with complete information and unlimited computational capacity), limiting their applicability in real-world settings, where agents face \textit{bounded rationality} (i.e., agents make decisions based on limited information and computational constraints) and dynamic uncertainties.

To address scalability and real-world uncertainties, recent studies have explored \textit{decentralized paradigms} that decompose coordination into localized decisions. These methods leverage agent-level computation and local communication to approximate Nash equilibria (NE) more efficiently. For example, \textit{Williams et al.}~\cite{williams2023distributed} used interaction graphs to cluster agents for NE search; \textit{Wang et al.}~\cite{wang2021game} applied iterative Best Response (iBR) for convergence refinement; and \textit{Cui et al.}~\cite{cui2024nash} adopted iBR in a Nash-Stackelberg racing framework. However, these works focus mainly on coordination and convergence, with limited treatment of \textit{safety} or \textit{robustness}, motivating the need for decentralized approaches that are both scalable and safety-aware under uncertainty.

\subsubsection{Safety-Aware Motion Planning under Uncertainty}

In the domain of safety-aware motion planning, probabilistic approaches are often used to handle uncertainty through stochastic modeling. Recent methods \cite{lembono2021probabilistic, liu2023tight, xu2024quadcopter} focus on uncertainty propagation in dynamic systems to generate risk-sensitive trajectories for single-agent scenarios, incorporating probabilistic representations of disturbances and model errors for formal safety guarantees. However, extending these methods to multi-agent systems remains challenging due to the computational burden of propagating uncertainties across coupled agents.

Another widely adopted strategy is Hamilton-Jacobi (HJ) reachability analysis, which computes forward reachable sets (FRSs) --- the set of all possible states an agent can reach under bounded disturbances --- to enhance robustness in uncertain environments \cite{bansal2017hamilton}. For example, \textit{Seo et al.} used ellipsoidal bounds for error state FRSs, enabling real-time uncertainty-aware trajectory optimization for collision avoidance \cite{seo2019robust,seo2022real}. While effective in single-agent settings, these methods face challenges in multi-agent systems due to inter-agent coupling. Specifically, the complexity of jointly evaluating reachable sets across multiple agents limits the direct application of HJ reachability techniques in dense multi-agent settings.

Beyond traditional analytical methods, learning-based approaches have recently shown promise for safe multi-agent motion planning under uncertainty \cite{vinod2022safe, zhu2024multi, zhang2025gcbf+, zhou2025robust}. These methods employ \textit{data-driven models} to approximate intricate safety constraints or value functions, thereby enabling greater adaptability in unstructured or dynamic environments. Nonetheless, their effectiveness is strongly contingent upon the quality and representativeness of the training data. Moreover, establishing rigorous theoretical guarantees for safety remains challenging, which raises concerns regarding their dependability when deployed in safety-critical real-world systems.
\subsection{Novelty and Contribution}
In this work, we address \textit{multi-agent motion planning in dynamic and uncertain environments, focusing primarily on ensuring safety}. To this end, we propose the \textit{Reachability-Enhanced Dynamic Potential Game (RE-DPG)} framework, which integrates \textit{DPG formulation, neighborhood-based iterative strategy updates, and reachability-based safety mechanisms}. To the best of our knowledge, this work represents an initial attempt to integrate DPG with formal reachability-based safety guarantees for multi-agent motion planning under uncertainty, and to explore the distinctive properties of such integration. Our key contributions can be summarized as follows:

\begin{figure*}[htbp]
% \vspace{-6.5mm}
    \centering
    \includegraphics[width=.85\linewidth]{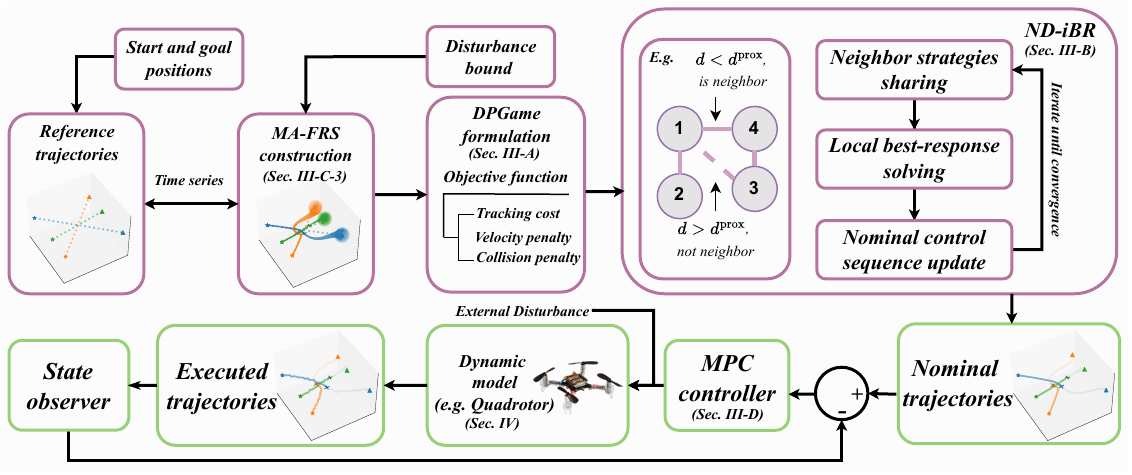}
    \caption{System overview of RE-DPG. Initial states, goal states, and neighborhood information are processed by the MA-FRS and DPG formulation modules, where safety-aware cost functions are established. The ND-iBR module then iteratively refines the planned nominal trajectories by incorporating neighboring agents' strategies. Note that the planning and tracking of nominal trajectories are decoupled. The finalized trajectories are executed using a receding horizon MPC controller to handle disturbances, while a state estimation mechanism continuously provides feedback to ensure safe trajectory execution in dynamic environments.}
    \label{fig:systemoverview}
\end{figure*}

\noindent
$\bullet$ We address the problem of safe multi-agent motion planning in dynamic and uncertain environments subject to disturbances by formulating a DPG that captures inter-agent interactions. Within this framework, each agent minimizes a cost function that balances task efficiency (e.g., timely goal attainment) and safety (e.g., collision avoidance). The objective is to compute the corresponding NE, which defines the coordinated behavior of all agents under this trade-off-driven cost structure.

\noindent
$\bullet$ We develop a \textit{Neighborhood-Dominated iterative Best Response (ND-iBR)} scheme, grounded in an iterated $\varepsilon$-BR (i$\varepsilon$-BR) process. This method ensures finite-step convergence to an $\varepsilon$-NE while enabling each agent to compute strategies based solely on local interactions, thereby achieving distributed and scalable NE approximation.

\noindent
$\bullet$ We further introduce a \textit{Multi-Agent Forward Reachable Set (MA-FRS)} mechanism to enhance safety in motion planning under bounded disturbances. By embedding \textit{MA-FRS-based penalties} into the agents' cost functions, the resulting framework proactively ensures that the planned agents' trajectories remain feasible while avoiding potential collisions among them in dynamic and uncertain environments.

\noindent
$\bullet$ We validate the RE-DPG framework through both simulations and real-world experiments. The results demonstrate that our approach consistently enables safe and robust trajectory planning across both 2D and 3D dynamic scenarios, signaling its effectiveness in practical multi-agent environments.
% \section{Overview and Key Modeling}
\section{Preliminaries}
\subsection{Notations and Definitions}
In this paper, subscripts are generally used to indicate agent indices while superscripts indicate time steps, labels, or matrix operations. Notation $(\cdot)^\mathsf{T}$ denotes the \textit{transpose} of a matrix and $\text{Tr}(\cdot)$ refers to the \textit{trace} operation, which computes the sum of the diagonal elements of a matrix. Additionally, $\Vert v \Vert$ represents the \textit{Euclidean 2-norm} of vector $v$.

\begin{definition}[Minkowski Sum]
\label{def:Minkowski_Sum}
For two sets $A$ and $B$ in Euclidean space, their Minkowski sum is the set of all vectors obtained by adding a vector from $A$ to that from $B$, given by:
\begin{equation}
\begin{aligned}
A\oplus B=\{x+y \mid x\in A, y\in B\}.
\end{aligned}
\end{equation}
\end{definition}

\begin{definition}[Ellipsoids in $n$-Dimensional Space]
\label{def:ellipsoid}
An ellipsoid $\mathcal{E}$ in $n$-dimensional Euclidean space is represented by the set:
\begin{equation}
\begin{aligned}
\mathcal{E} = \left\{ x \in \mathbb{R}^n \mid (x - c_0)^T Q^{-1} (x - c_0) \leq 1 \right\},
\end{aligned}
\end{equation}
where $c_0$ is the center and shape matrix $Q$ is positive definite. 
\end{definition}

\begin{definition}[Translation of an Ellipsoid]
\label{def:ellipsoid_translation}
Given an ellipsoid $\mathcal{E}$ with center $c_0$ and shape matrix $Q$, the \textit{translation} of $\mathcal{E}$ to a new center $c\neq c_0$ is defined as:
\begin{equation}
T_c(\mathcal{E}) := \left\{ x\in \mathbb{R}^n \mid (x - c)^T Q^{-1} (x - c) \leq 1 \right\}.
\end{equation}
\end{definition}

\begin{definition}[Ellipsoidal Approximation of the Minkowski Sum for Concentric Ellipsoids \cite{seo2019robust}]
\label{def:minkowski_sum_ellipsoid}
The Minkowski sum of multiple ellipsoids is generally not an ellipsoid and lacks a closed-form expression; in the concentric case, it can be approximated by a single ellipsoid for tractability.  For $n$ concentric ellipsoids (i.e., all centered at the origin) with shape matrices $\{Q_i\}_{i=1}^{n}$, the shape matrix of the ellipsoidal approximation for their Minkowski sum is defined as:
\begin{equation}
\begin{aligned}
\mathop{\boxplus}\limits_{i=1}^n Q_i &=Q_1\boxplus\ldots\boxplus Q_n\\
&= \left(\sum_{i=1}^{n} \sqrt{\operatorname{Tr}\left(Q_i\right)}\right)\left(\sum_{i=1}^{n} \frac{Q_i}{\sqrt{\operatorname{Tr}\left(Q_i\right)}}\right),
\end{aligned}
\end{equation}
which approximates the Minkowski sum of multiple concentric ellipsoids by a single ellipsoid whose shape matrix is designed to tightly enclose the set corresponding to the true Minkowski sum through minimizing the volume of this enclosing ellipsoid.
\end{definition}

% The Minkowski sum of two sets \( A \) and \( B \) is defined as the set of all points that can be formed by adding each element of \( A \) to each element of \( B \). The \textit{Minkowski sum} of two ellipsoidal shape matrices \( Q_1 \) and \( Q_2 \), denoted as \( Q_1 \oplus Q_2 \), is defined as the resulting shape matrix of the Minkowski sum of the two ellipsoids. 
% In a nutshell, multi-agent motion planning centers on developing optimal control strategies for multiple agents operating collaboratively within a shared environment. A key challenge in this process lies in accurately modeling the dynamic interactions among agents, as these interactions directly influence individual agent behaviors. The specific motion planning problem addressed in this work involves generating control inputs that guide agents efficiently from their initial to target states, while ensuring both time efficiency and robustness to achieve collision-free navigation in dynamic settings.
% \vspace{-1em}
\subsection{Overview and Key Modeling Components}
In a nutshell, multi-agent motion planning centers on developing optimal control strategies for multiple agents operating collaboratively within a shared environment. The specific motion planning problem addressed in this work involves generating control inputs that guide agents from their initial to target states, while ensuring both time efficiency and robustness to achieve collision-free navigation in dynamic settings.

Formally, we consider a multi-agent system that consists of a set of agents $\mathcal{N} = \{1, \ldots, N\}$. Each agent $i \in \mathcal{N}$ operates within its state space $\mathcal{X}_i \subseteq \mathbb{R}^{n_x}$ (e.g., agent's position, velocity, and orientation) and control input space $\mathcal{U}_i \subseteq \mathbb{R}^{n_u}$ (e.g., acceleration commands, steering angles, or thrust forces applied to the agent), where $n_x$ and $n_u$ are the respective dimensions of these spaces. The joint state space and control input space across the agents are denoted by $\mathcal{X} = \mathcal{X}_1 \times \cdots \times \mathcal{X}_N \subseteq \mathbb{R}^{Nn_x}$ and $\mathcal{U} = \mathcal{U}_1 \times \cdots \times \mathcal{U}_N \subseteq \mathbb{R}^{Nn_u}$, respectively. At each discrete time step $t$, the state vector of agent $i$ is denoted by $x_i^{(t)}\in\mathcal{X}_i$, and its control input by $u_i^{(t)}\in\mathcal{U}_i$. Accordingly, the concatenated states and control inputs of all agents at time $t$ are represented as $\boldsymbol{x}^{(t)}=(x_1^{(t)},\cdots, x_N^{(t)})\in\mathcal{X}$ and $\boldsymbol{u}^{(t)}=(u_1^{(t)},\cdots, u_N^{(t)})\in\mathcal{U}$. For each agent, the objective is to determine a proper sequence of control inputs $\mathbf{u}_i = \{u_i^{(t)}\}_{t=0}^{T-1}$ that guides the agent from an initial state $x_i^{(0)}$ to a final/goal state $x_i^\text{F}$ over a finite time horizon $T$. We denote the collective states and control inputs of all agents except agent $i$ at time $t$ as $\boldsymbol{x}_{-i}^{(t)}$ and $\boldsymbol{u}_{-i}^{(t)}$, where we similarly define the collection $\mathbf{u}_{-i}=(\mathbf{u}_1,\ldots,\mathbf{u}_{i-1},\mathbf{u}_{i+1},\ldots,\mathbf{u}_N)$.\footnote{Note that the difference between \( \boldsymbol{u}^{(t)} \) and \( \mathbf{u}_i \) lies in their definitions: \( \boldsymbol{u}^{(t)} \) represents the control inputs of all agents at a single time step, while \( \mathbf{u}_i \) represents the control inputs of a single agent over the planning horizon. Therefore, $\{\boldsymbol{u}^{(t)}\}^{T-1}_{t=0}$ and $\{\mathbf{u}_i\}_{i=1}^N$ are equivalent.} Note that the control input \( u_i^{(t)} \) directly determines the evolution of agent $i$'s state \( x_i^{(t+1)} \) as described by the following state transition equation:
\begin{equation}
\begin{aligned}
{x}_i^{(t+1)} = f_i\left({x}_i^{(t)}, {u}_i^{(t)}\right),\quad t \in &\{0,\ldots,T-1\}, 
\end{aligned}
\label{eq:sysdyna}
\end{equation}
where $f_i(\cdot)$ represents the state transition function that maps the current state and control input of agent $i$ to its next state.
% \subsection*{Preliminaries}
% In this paper, subscripts are generally used to indicate agent indices while superscripts indicate time steps, labels, or matrix operations. $(\cdot)^\mathsf{T}$ denotes the \textit{transpose} of a matrix. Additionally, $\text{Tr}(\cdot)$ refers to the \textit{trace} operation, which computes the sum of the diagonal elements of a matrix. The Minkowski sum of two sets \( A \) and \( B \) is defined as the set of all points that can be formed by adding each element of \( A \) to each element of \( B \). The \textit{Minkowski sum} of two ellipsoidal shape matrices \( Q \) and \( Q \), denoted as \( A \oplus B \), is defined as the resulting shape matrix of the Minkowski sum of the two ellipsoids. 

\section{Safe Multi-Agent (MA) Motion Planning via Integrating Forward Reachable Set (FRS) into Dynamic Potential Game (DPG)}
In this section, we present a solution for multi-agent motion planning, focusing on safety and robustness through a hybrid approach combining the DPG framework with an MA-FRS mechanism. We begin by formulating a DPG-based motion planning problem to determine safe control inputs over a finite time horizon, considering agent interactions (Sec.~\ref{Sec:DPG}). To ensure scalability, we then propose the ND-iBR scheme, which decomposes the problem into distributed subproblems, allowing each agent to iteratively compute its strategy and approximate the NE in a decentralized manner (Sec.~\ref{Sec:iBR}). Afterward, given the uncertainties in the scenario, such as external disturbances and dynamic environmental variations, we focus on modeling uncertainty propagation and inter-agent interactions. To this end, we incorporate MA-FRS-based penalties to enable uncertainty-aware trajectory planning and proactive collision avoidance (Sec.~\ref{Sec:Reachset}), and then demonstrate an approach to reach the NE in our setting of interest (Sec.~\ref{Sec:NE}). An overview of our proposed framework is illustrated by Fig. \ref{fig:systemoverview}. 

\subsection{DPG Formulation for Multi-Agent Motion Planning}\label{Sec:DPG}

To systematically model the multi-agent motion planning problem, we follow a standard approach \cite{zazo2016dynamic}. It is worth mentioning that our major contribution is not the subsequent standard formulations but rather in the solution design in Secs.~\ref{Sec:iBR}, ~\ref{Sec:Reachset}, and \ref{Sec:NE}. To enable performance quantification, we first define a \textit{performance criterion}. Specifically, we formulate a \textit{cost function} \( J_i \) for each agent $i$ as follows:
\begin{subequations}
\begin{align}
J_i(\boldsymbol{x}^{(0)},(\mathbf{u}_{i}, {\mathbf{u}}_{-i})) = \sum^{T-1}_{t=0} L_i(\boldsymbol{x}^{(t)},{u}^{(t)}_i) + L^\text{F}_i(\boldsymbol{x}^{(T)}),
\label{eq:Ji}
\end{align}
\end{subequations}
\noindent\hspace{-0.5em}{where} $L_i(\cdot)$ and $L^\text{F}_i(\cdot)$ represent the \textit{stage cost} and \textit{terminal cost} of agent $i$, respectively. To ensure safe and efficient multi-agent motion planning, we consider a decomposed formulation for $L_i(\cdot)$ and $L^\text{F}_i(\cdot)$ as follows:
\begin{subequations}
\begin{align}
\label{Li_c}
L_i(\boldsymbol{x}^{(t)},{u}^{(t)}_i)&=C_i^{\text{Tr}}(x^{(t)}_i,u^{(t)}_i)+\sum_{i\neq j,j\in\mathcal{N}}C_{i,j}^{\text{Coup}}(x^{(t)}_i, x^{(t)}_j), \\
\label{Lif_c}
L_i^\text{F}(\boldsymbol{x}^{(T)})&=C_i^{\text{Tr,F}}(x^{(T)}_i)+\sum_{i\neq j,j\in\mathcal{N}}C_{i,j}^{\text{Coup}}(x^{(T)}_i, x^{(T)}_j), 
\end{align}
\end{subequations}
\noindent where $C_i^{\text{Tr}}(\cdot)$ represents the tracking cost, which depends solely on agent 
$i$'s own state and control input, capturing objectives such as goal reaching and control effort, while $C_{i,j}^{\text{Coup}}(\cdot)$ represents the coupling cost between agent $i$ and $j$, modeling interactive effects such as collision avoidance.  We assume that the coupling cost is symmetric, meaning $C_{i,j}^{\text{Coup}}(x^{(t)}_i, x^{(t)}_j)=C_{j,i}^{\text{Coup}}(x^{(t)}_j, x^{(t)}_i), \forall i \neq j, i,j\in \mathcal{N}$. The detailed design of these costs will be provided in Sec.~\ref{Sec:Reachset}.

Next, we introduce the concept of a \textit{strategy}, which defines how each agent selects its motion over time to minimize its cost. In this context, we adopt an \textit{open-loop strategy}~\cite{bhatt2023efficient}, where each agent precomputes a time-dependent control policy based on the initial system configuration, including its own initial state and the anticipated decisions of other agents. Specifically, for each agent $i$, the strategy is represented as a mapping $\mathbf{u}_i(\boldsymbol{x}^{(0)}, t) := u_i^{(t)}$, which remains fixed during execution and yields time-dependent control inputs. Although expressed in an open-loop form, this representation serves mainly as a modeling abstraction, while our framework later incorporates mechanisms that approximate closed-loop behavior for improved robustness. The open-loop NE under these strategies is defined below.

\begin{definition}[Open-Loop Nash equilibrium (NE)]
\label{def:open_loop_NE}
In a game with players $\mathcal{N}$, state space $\mathcal{X}$, control input space $\mathcal{U}$, and cost functions $\{J_i(\cdot,\cdot)\}_{i\in\mathcal{N}}$, a control input profile $\mathbf{u}^* = (\mathbf{u}_i^*, \mathbf{u}_{-i}^*)$ is an open-loop NE if:
\begin{equation}\label{eq:NEdef}
J_i(\boldsymbol{x}^{(0)}, (\mathbf{u}_i^*, \mathbf{u}_{-i}^*)) \leq J_i(\boldsymbol{x}^{(0)}, (\mathbf{u}_i, \mathbf{u}_{-i}^*)), \quad \forall \mathbf{u}_i.
\end{equation}
\end{definition}
This ensures no agent can reduce its cost by unilaterally changing its strategy while others stick to their equilibrium strategies. Direct computation of the NE is challenging due to the $N$ coupled constrained optimal control problems. However, the multi-agent motion planning problem can be reformulated as a DPG, simplifying NE computation through a single multivariate optimization problem.

% \vspace{-1.25mm}
\begin{definition}[DPG for Multi-Agent Motion Planning]
\label{def:DPG}
A game is a DPG if the stage cost $L_i(\boldsymbol{x}^{(t)}, \mathbf{u}_i^{(t)})$ and terminal cost $L_i^\text{F}(\boldsymbol{x}^{(T)})$ for each agent $i$ can be decomposed as:
\begin{subequations}
\begin{align}
L_i(\boldsymbol{x}^{(t)}, \mathbf{u}_i^{(t)}) &= P(\boldsymbol{x}^{(t)}, \boldsymbol{u}^{(t)}) + \Theta_i(\boldsymbol{x}_{-i}^{(t)}, \mathbf{u}_{-i}^{(t)}), \\
L_i^\text{F}(\boldsymbol{x}^{(T)}) &= R(\boldsymbol{x}^{(T)}) + \Xi_i(\boldsymbol{x}_{-i}^{(T)}),
\end{align}
\label{eq:Li_and_Lif_decompose}
\end{subequations}
\hspace{-0.25em}where $P(\cdot, \cdot)$ and $R(\cdot)$ are global potential functions depending on the joint states and control inputs of all agents, while $\Theta_i(\cdot, \cdot)$ and $\Xi_i(\cdot)$ depend only on the states and inputs of other agents.
\end{definition}
% \vspace{-0.75mm}
% From \eqref{Li_poten}-\eqref{Lif_poten}, we can see that functions \( P(\cdot,\cdot) \) and \( R(\cdot) \) can capture gradient information (and thus the changes) in $L_i(\cdot)$ and $L^\text{F}_i(\cdot)$, and consequently also in \( J_i(\cdot) \), for which we call them potential functions (PFs). We can define a feasible detailed calculation of these PFs as following:
% \begin{subequations}
% \begin{align}
%     \label{P}
%     P(\boldsymbol{x}^{(t)}, \boldsymbol{u}^{(t)}) &= \mathop{\sum}_{i \in \mathcal{N}} L_i(x^{(t)}_{i},\boldsymbol{u}^{(t)}), \\
%     \label{R}
% R(\boldsymbol{x}^{(T)}) &= \mathop{\sum}_{i \in \mathcal{N}}  L^\text{F}_i(x^{(T)}_{i}).
% \end{align}
% \end{subequations}

% Combining \eqref{Li_poten}-\eqref{Lif_poten} and \eqref{P}-\eqref{R}, we can obtain:
% \begin{subequations}
% \begin{align}
%     \label{Theta}
%     \Theta_i({x}_{-i}^{(t)}, {u}_{-i}^{(t)}) &= -\mathop{\sum}_{j\neq i, j \in \mathcal{N}} L_j(x^{(t)}_{j},\boldsymbol{u}^{(t)}), \\
%     \label{Xi}
% \Xi_i({x}_{-i}^{(t)}) &= -\mathop{\sum}_{j\neq i, j \in \mathcal{N}}  L^\text{F}_j(x^{(T)}_{j}).
% \end{align}
% \end{subequations}
Under the cost structure in \eqref{Li_c}-\eqref{Lif_c}, with the symmetric condition on $C_{i,j}^{\text{Coup}}(\cdot)$, our problem setup can be transformed to a DPG with the following PFs:
\begin{subequations}
\begin{align}
\label{P_define}
\hspace{-3mm}P(\boldsymbol{x}^{(t)}, \boldsymbol{u}^{(t)})&=\sum_{i\in\mathcal{N}}C_i^{\text{Tr}}(x^{(t)}_i,u^{(t)}_i)+\sum_{\substack{i< j\\i,j\in\mathcal{N}}}C_{i,j}^{\text{Coup}}(x^{(t)}_i, x^{(t)}_j),\raisetag{0.5em}\hspace{-3mm} \\
\label{R_define}
R(\boldsymbol{x}^{(T)})&=\sum_{i\in\mathcal{N}}C_i^{\text{Tr,F}}(x^{(T)}_i)+\sum_{\substack{i< j\\i,j\in\mathcal{N}}}C_{i,j}^{\text{Coup}}(x^{(T)}_i, x^{(T)}_j).\raisetag{1.0em}
\end{align}
\end{subequations}
\noindent Note that \(P(\cdot,\cdot)\) and \(R(\cdot)\) are not simple sums of individual agent costs. Specifically, while tracking costs aggregate across all agents, each coupling term \(C_{i,j}^{\text{Coup}}(\cdot)\) appears only once for unordered pairs \((i,j)\) with \(i<j\) to avoid double-counting. These  PFs yield the following forms for \(\Theta_i(\cdot,\cdot)\) and \(\Xi_i(\cdot)\):
% \noindent It is important to note that \(P(\cdot,\cdot)\) and \(R(\cdot)\) are \textit{not} simple summations of individual agent costs. In particular, while the tracking costs are aggregated across all agents, each coupling cost term \(C_{i,j}^{\text{Coup}}(\cdot)\) is included \textit{only once} for each unordered agent pair $(i,j)$ where condition \(i<j\) is induced to avoid double-counting. These structured PFs lead to the following forms of $\Theta_i(\cdot,\cdot)$ and $\Xi_i(\cdot)$:
\begin{subequations}
\begin{align}
&\hspace{-4mm}\Theta_i(\boldsymbol{x}_{-i}^{(t)}, \boldsymbol{u}_{-i}^{(t)}) = \notag\\
- &\sum_{\substack{j\in\mathcal{N} \\ j\neq i}} C_j^\text{Tr}(x_j^{(t)}, u_j^{(t)}) 
- \sum_{\substack{j\in\mathcal{N} \\ j\neq i}} \sum_{\substack{k\in\mathcal{N} \\ k>j,\, k\neq i}} C_{j,k}^\text{Coup}(x_j^{(t)}, x_k^{(t)}),\\
&\hspace{-4mm}\Xi_i(\boldsymbol{x}_{-i}^{(T)}) =
{-} \sum_{\substack{j\in\mathcal{N} \\ j\neq i}} C_j^\text{Tr,F}(x_j^{(T)}) 
{-} \sum_{\substack{j\in\mathcal{N} \\ j\neq i}} \sum_{\substack{k\in\mathcal{N} \\ k>j,\, k\neq i}} C_{j,k}^\text{Coup}(x_j^{(T)}, x_k^{(T)}).\hspace{-2mm}
\end{align}
\end{subequations}
It can be observed that the above-defined PFs \(P(\cdot,\cdot)\) and \(R(\cdot)\) depend on the states and control inputs of all agents, whereas \(\Theta_i(\cdot,\cdot)\) and \(\Xi_i(\cdot)\) are independent of agent \(i\)'s own state and control input, satisfying the conditions in Definition \ref{def:DPG}.

To explicitly compute the NE of the above-defined DPG, we can formulate the problem as a centralized optimization formulation that minimizes the cumulative sum of \(P(\cdot,\cdot)\) over the planning horizon, along with the terminal cost \(R(\cdot)\), as also shown in \cite{zazo2016dynamic} as follows:
\begin{equation*}
\begin{aligned}
&\hspace{-1mm}\boldsymbol{\mathcal{P}}: \mathop{\min}_{\boldsymbol{u}^{(0)},\ldots,\boldsymbol{u}^{(T-1)}} \quad  \sum_{t=0}^{T-1} P(\boldsymbol{x}^{(t)}, \boldsymbol{u}^{(t)}) + R(\boldsymbol{x}^{(T)}) \\
&\hspace{-1mm}\text{s.t.}~~~  {x}_i^{(t+1)} = {f_i}({x_i^{(t)}}, {u}_i^{(t)}),~
 t \in \{0,\ldots,T-1\},~\forall i \in \mathcal{N}.~~ \text{{(C1)}} 
% &\quad\quad\quad\quad\quad.
\end{aligned}
\end{equation*}

% \vspace{-3mm}
\subsection{ND-iBR-based Problem Transformation}\label{Sec:iBR}
Prior methods for solving the problem \( \boldsymbol{\mathcal{P}} \) often rely on centralized frameworks \cite{bhatt2023efficient, zhang2024game, bhatt2025strategic}, calling for full-state sharing and scale poorly under dense agent coupling. 
To mitigate the above limitations, we resort to an iterative best-response (iBR) approach as a scalable solution. Yet, in continuous control scenarios, conventional iBR is not guaranteed to converge to an exact NE \cite{zanardi2021urban}. To remedy this, we adopt an iterated $\varepsilon$-BR (i$\varepsilon$-BR) mechanism, where agents revise their strategies only if doing so yields at least an $\varepsilon$-level cost reduction. This relaxation ensures that the process converges, within finite iterations, to an approximate NE ($\varepsilon$-NE).
% To overcome these limitations, we adopt an iterative best-response (iBR) approach for scalable computation. However, in continuous control settings, classical iBR may fail to guarantee convergence to an exact NE \cite{zanardi2021urban}. To address this, we introduce the iterated $\varepsilon$-best response (i$\varepsilon$-BR), which ensures each agent updates its strategy with at least $\varepsilon$ cost improvement, allowing convergence to an approximate Nash equilibrium ($\varepsilon$-NE) in finite iterations. 
This balances accuracy, safety, and computational efficiency. Hereafter, we present a theorem to formalize this process and note that the superscript $k$ (e.g., $\mathbf{u}^k$) denotes the index of the iBR iteration, which is distinct from the time index $t$ used in $\boldsymbol{x}^{(t)}$ or $\boldsymbol{u}^{(t)}$:
% \vspace{-0.5em}
\begin{theorem}[Convergence of i$\varepsilon$-BR]\label{theorem:ibr_dpg}
Consider the above DPG with a set of agents $\mathcal N$. 
For each player $i\in\mathcal N$, let their strategy set be nonempty and compact. 
Suppose $J_i$ admits the decomposition \eqref{eq:Ji} where \( L_i(\cdot), L_i^\text{F}(\cdot)\) are both continuous, and \eqref{eq:Li_and_Lif_decompose} also holds.
Fix $\varepsilon>0$ and consider the iterated $\varepsilon$-best response (i$\varepsilon$-BR): 
at iteration $k$, choose any agent $i$ such that
\[
J_i(\boldsymbol{x}^{(0)},(\mathbf u_i^{k},\mathbf u_{-i}^k))-\inf_{\mathbf v_i} J_i(\boldsymbol{x}^{(0)},(\mathbf v_i,\mathbf u_{-i}^k))~\ge~\varepsilon,
\]
and set $\mathbf u_i^{k+1}$ as:
$\mathbf u_i^{k+1}=\arg\min_{\mathbf u_i} J_i\big(\boldsymbol{x}^{(0)}, \big(\mathbf u_i,\mathbf u_{-i}^{k}\big)\big)$
; if no such agent exists, stop. 
This process terminates in a finite number of iterations at a profile $\mathbf u^\varepsilon$ that is an $\varepsilon$-Nash equilibrium, i.e.,
\[
J_i(\boldsymbol{x}^{(0)},(\mathbf u_i^\varepsilon,\mathbf u_{-i}^\varepsilon))~\le~
\inf_{\mathbf v_i} J_i(\boldsymbol{x}^{(0)},(\mathbf v_i,\mathbf u_{-i}^\varepsilon))+\varepsilon,\quad \forall i\in\mathcal N.
\]
\end{theorem}
% \vspace{-0.5em}
\noindent\textit{Proof.}
Given fixed $k$ and $i$, we consider updating only the strategy $\mathbf u_i^k$ of agent $i$ while keeping $\mathbf u_{-i}^k$ unchanged.
By definition of the best response, we have
\(
J_i(\boldsymbol{x}^{(0)},(\mathbf u_i^{k},\mathbf u_{-i}^k)) - J_i(\boldsymbol{x}^{(0)},(\mathbf u_i^{k+1},\mathbf u_{-i}^k)) \;\ge\; 0.
\)
Now recall from \eqref{eq:Ji} and \eqref{eq:Li_and_Lif_decompose} that each cost admits the decomposition
\(
J_i(\mathbf u_i^k,\mathbf u_{-i}^k) = \Phi(\mathbf u_i^k,\mathbf u_{-i}^k) + \Psi_i(\mathbf u_{-i}^k),
\)
where the potential function is defined by
\(
\Phi(\mathbf u_i,\mathbf u_{-i}) := \sum_{t=0}^{T-1} P(\boldsymbol{x}^{(t)}, \boldsymbol{u}^{(t)}) + R(\boldsymbol{x}^{(T)}),
\)
and the residual $\Psi_i$ is independent of $\mathbf u_i$. 
Therefore, when only $\mathbf u_i$ is updated, we obtain
\(
J_i(\mathbf u_i^k,\mathbf u_{-i}^k) - J_i(\mathbf u_i^{k+1},\mathbf u_{-i}^k)
= \Phi(\mathbf u_i^k,\mathbf u_{-i}^k) - \Phi(\mathbf u_i^{k+1},\mathbf u_{-i}^k).
\)
Thus the potential function $\Phi(\cdot, \cdot)$ is non-increasing across iBR updates. 
Under the i$\varepsilon$-BR rule, $\Phi(\cdot, \cdot)$ decreases by at least $\varepsilon$ at each update. 
Since $\Phi(\cdot, \cdot)$ is continuous and the strategy set is compact, $\Phi(\cdot, \cdot)$ has a minimum value \(\Phi_{\min}\). 
Therefore, the total number of updates is bounded above by $(\Phi(\mathbf u_i^0,\mathbf u_{-i}^0)-\Phi_{\min})/\varepsilon$, and the process must terminate in finitely many steps. 
At termination, no player can reduce its cost by $\varepsilon$ or more, which is exactly the condition of an $\varepsilon$-Nash equilibrium.
\hfill$\blacksquare$

% Given fixed $k$ and $i$, we consider updating only the strategy $\mathbf u_i^k$ of agent $i$ while keeping $\mathbf u_{-i}^k$ unchanged.
% By definition of the best response, we have \(J_i(\boldsymbol{x}^{(0)},(\mathbf u_i^{k},\mathbf u_{-i}^k)) - J_i(\boldsymbol{x}^{(0)},(\mathbf u_i^{k+1},\mathbf u_{-i}^k)) \ge 0.
% \)
% Using the decomposition \eqref{eq:Ji} and the fact that the residual terms in \eqref{eq:Ji} do not depend on $\mathbf u_i$, the same inequality holds for the potential terms $\sum_{t=0}^{T-1} P(\boldsymbol{x}^{(t)}, \boldsymbol{u}^{(t)}) + R(\boldsymbol{x}^{(T)})$; hence these terms are non-increasing across iBR updates. 
% Now, under i$\varepsilon$-BR, the selection rule ensures
% \(
% J_i(\boldsymbol{x}^{(0)},(\mathbf u_i^{k},\mathbf u_{-i}^k))-
% J_i(\boldsymbol{x}^{(0)},(\mathbf u_i^{k+1},\mathbf u_{-i}^k))~\ge~\varepsilon,
% \)
% and thus the potential value $\sum_{t=0}^{T-1} P(\boldsymbol{x}^{(t)}, \boldsymbol{u}^{(t)}) + R(\boldsymbol{x}^{(T)})$ decreases by at least $\varepsilon$ at each update. 
% According to \eqref{eq:Li_and_Lif_decompose} and compactness of the feasible set, this potential is bounded below; hence only finitely many updates are possible. 
% Upon termination, no player can improve its cost by $\varepsilon$ or more, which can be exactly seen as the stated $\varepsilon$-NE condition. 
% \hfill$\blacksquare$
% \vspace{-0.5em}
\begin{remark}[Practical sufficiency of $\varepsilon$-NE and local minimum]
The convergence of $\varepsilon$-NE to an exact NE (i.e., when $\varepsilon=0$) is generally not guaranteed. Nevertheless, fixing a small tolerance and terminating i$\varepsilon$-BR at an $\varepsilon$-NE is typically sufficient in practice \cite{zanardi2021urban}. 
Additionally, the potential function $\Phi(\cdot, \cdot)$ may admit multiple local minima; the particular one reached by our method depends on the order in which agents update. Obtaining the global minimum would require enumerating all possible update sequences and comparing their values of $\Phi(\cdot, \cdot)$, which computationally expensive. We therefore accept local minimum of $\Phi(\cdot, \cdot)$ under a fixed update rule (e.g., ordered by agent index), and as long as the resulting NE avoids collisions, it suffices for safe multi-agent planning.
\end{remark}
% \vspace{-0.5em}
The preceding analysis assumes global state access, which is often unrealistic in practice. In real-world scenarios, agents rely on local observations from nearby neighbors. We formalize neighborhoods by defining agents \(i, j \in \mathcal{N}\) as neighbors at time \(t\) if their distance \(d^{(t)}_{i,j}\) is less than a proximity threshold \(d^{\mathrm{Prox}} \geq 2v^{\mathrm{Max}}\Delta t\), representing the worst-case one-step closing distance. The set of neighbors for agent $i$ is then $\tilde{\mathcal{N}}^{(t)}_i = \{j~|~d^{(t)}_{i,j}<d^{\mathrm{Prox}}, j\neq i, j\in \mathcal{N}\}$.
Given this definition of neighborhood, we introduce the Neighborhood-Dominated iterative Best Response (ND-iBR) scheme, where each agent updates its strategy based on the fixed strategies of its local neighbors. Note that while the procedure follows the i$\varepsilon$-BR framework, we retain ND-iBR as the name to highlight its reliance on local interactions. The localized cost function considering only neighbors' influence is defined as $\tilde{J_i}(\cdot,\cdot)$, and the ND-iBR update rule is modified to the following i$\varepsilon$-BR-based optimization:
% \vspace{-0em}
\begin{equation}
\begin{aligned}
\mspace{-20mu}{\boldsymbol{\tilde{\mathcal{P}}}}^k_i: \quad \mathbf{u}_i^{k+1} = \arg\min_{\mathbf{u}_i} \; \tilde{J}_i\left( \boldsymbol{x}^{(0)}, \left( \mathbf{u}_i, \tilde{\mathbf{u}}^k_{-i} \right) \right) \\
\text{s.t.} \quad \text{{(C1)}}, \tilde{J}_i\left( \boldsymbol{x}^{(0)}, \left( \mathbf{u}_i, \tilde{\mathbf{u}}^k_{-i} \right) \right) - \inf_{\mathbf{v}_i} \tilde{J}_i\left( \boldsymbol{x}^{(0)}, \left( \mathbf{v}_i, \tilde{\mathbf{u}}^k_{-i} \right) \right) \geq \varepsilon.
\end{aligned}
\label{eq:NDiBRPi}
\end{equation}
\begin{remark}[Neighbor screening and computational benefits]
The proximity constraint $d^{\mathrm{Prox}} \ge 2v^{\mathrm{Max}}\Delta t$ serves as a one-step conservative screening: agents beyond this bound cannot reach agent $i$ within the next interval, even under worst-case closing speeds with different accelerations or turning rates. Also, neighbor sets $\tilde{\mathcal N}_i^{(t)}$ are recomputed at every planning step from the latest nominal predictions, reflecting agents entering or leaving the $d^{\mathrm{Prox}}$ threshold. This neighborhood-based approach reduces the number of coupling terms per agent, leading to smaller optimization subproblems without unduly sacrificing coordination performance. Moreover, the coupling term $C_{i,j}^{\mathop{\mathrm{Coup}}}$ will be redesigned in Sec.~\ref{Sec:Reachset} to reflect collision-related costs, and the exclusion of non-neighboring agents introduces only negligible approximation error in practice. Consequently, $\tilde{J}_i(\cdot,\cdot)$ retains the boundedness and monotonicity properties required by Theorem~\ref{theorem:ibr_dpg}, thereby justifying the use of ND-iBR for computing the $\varepsilon$-NE in a distributed manner.
\end{remark}

% \vspace{-1.5em}
\subsection{Cost Enhancement via Reachability Analysis}\label{Sec:Reachset}
% \vspace{-0em}
Building on the decentralized framework in Sec.~\ref{Sec:iBR}, we enhance the cost function by incorporating safety through MA-FRS. A key innovation is the introduction of MA-FRSs, which capture uncertainty propagation and dynamic interactions among multiple agents, a largely unexplored aspect in the literature. We add tracking cost and velocity constraints to ensure goal-directed behavior and dynamic limitations. This formulation maintains compatibility with DPG properties while integrating both goal-oriented objectives and interaction-aware constraints for safe multi-agent motion planning. We revise the cost function components in \eqref{eq:Ji} as follows:
% Building on the decentralized framework introduced in Sec.~\ref{Sec:iBR}, we now enhance the cost function to incorporate safety via MA-FRS. Here, a novel aspect of our approach is the introduction of MA-FRSs, which explicitly capture uncertainty propagation and dynamic interactions among multiple agents - an underexplored aspect in the literature. First, we introduce tracking cost and velocity constraints to ensure goal-directed behavior and dynamic limitations.  Our formulation ensures that the resulting optimization problem remains fully compatible with the properties of a DPG, while simultaneously capturing both \textit{goal-oriented objectives} and \textit{interaction-aware constraints} essential for safe multi-agent motion planning. We revise the individual components of the cost function in \eqref{eq:Ji} as follows:
\begin{equation}
\label{eq:tilde_Ji}
\begin{aligned}
\tilde{J_i}(\boldsymbol{x}^{(0)},(\mathbf{u}_{i}, \tilde{\mathbf{u}}_{-i})) = 
\sum^{T-1}_{t=0} \tilde{L_i}(\boldsymbol{x}^{(t)},{u}^{(t)}_i) + \tilde{L}^\text{F}_i(\boldsymbol{x}^{(T)}),
\end{aligned}
\end{equation}
where we use the following neighborhood-based definitions:
\begin{subequations}
\begin{align}
\tilde{L}_{i}(\boldsymbol{x}^{(t)},{u}^{(t)}_i) &= c_i^{(t)} + c_{i}^{\text{V},(t)} + \sum_{j\in\tilde{\mathcal{N}}^{(t)}_i} c_{i,j}^{\text{FRS},(t)}, \label{eq:LiLftilde1}\\
\tilde{L}^\text{F}_{i}(\boldsymbol{x}^{(T)}) &= c^\text{F}_i + c_{i}^{\text{V},(T)} + \sum_{j\in\tilde{\mathcal{N}}^{(t)}_i}c_{i,j}^{\text{FRS},(T)}\label{eq:LiLftilde2}.
\end{align}
\end{subequations}
We next describe the terms used in \eqref{eq:LiLftilde1}-\eqref{eq:LiLftilde2}. 
\noindent\subsubsection{Tracking Cost}
In \eqref{eq:LiLftilde1}-\eqref{eq:LiLftilde2}, we have decomposed the tracking cost in \eqref{Li_c} to stage and terminal costs $c_i^{(t)}$~and~$c_i^\text{F}$, which together penalize deviations from the reference trajectory and control effort, which are formally defined as:
\begin{subequations}
\begin{align}
c^{(t)}_i &= (x^{(t)}_i - r^{(t)}_i)^\mathsf{T} \boldsymbol{Q}_i (x^{(t)}_i - r^{(t)}_i) + (u^{(t)}_i)^\mathsf{T} \boldsymbol{R}_i (u^{(t)}_i), \label{eq:trackingcost1}\\
c^\text{F}_i &= (x^{(T)}_i - r^{\text{F}}_i)^\mathsf{T} \boldsymbol{Q}^\text{F}_i (x^{(T)}_i - r^{\text{F}}_i),\label{eq:trackingcost2}
\end{align}
\end{subequations}
\noindent where \( \boldsymbol{Q}_i, \boldsymbol{Q}^\text{F}_i \in \mathbb{R}^{n_x \times n_x}\) with \( \boldsymbol{Q}_i, \boldsymbol{Q}^\text{F}_i \succeq 0 \), and \( \boldsymbol{R}_i \in \mathbb{R}^{n_u \times n_u}\) with \( \boldsymbol{R}_i \succeq 0 \) are positive semi-definite weighting matrices. These terms penalize deviations of the agent’s state \( x_i^{(t)} \) from the reference trajectory \( r^{(t)}_i \) and the control effort \( u_i^{(t)} \). We presume that the reference trajectory \( r^{(t)}_i \) is generated by discretizing the line connecting the agent’s initial state to its goal state over the planning horizon, with time intervals of $\Delta t$. 
\noindent\subsubsection{Velocity Constraint Penalty} 
The velocity constraint penalty \( c_{i}^{\text{V},(t)} \) in \eqref{eq:LiLftilde1}-\eqref{eq:LiLftilde2} is introduced to discourage agents from exceeding a predefined speed limit \( v^{\mathrm{Max}} \), defined as:
\begin{equation}
\begin{aligned}
c_{i}^{\text{V},(t)} = 
\begin{cases}
\infty, & v^{\mathrm{Max}} - \|v_i^{(t)}\| \leq 0 , \\
0, & v^{\mathrm{Max}} - \|v_i^{(t)}\| > 0.
\end{cases}
\label{eq:cvti}
\end{aligned}
\end{equation}
where $v_i^{(t)}$  denotes the velocity of agent $i$ at time step $t$. However, the non-differentiability of the penalty function in \eqref{eq:cvti} pose challenges for solving the optimization problem defined in \eqref{eq:NDiBRPi} using standard gradient-based optimizers. To address this, we propose a \textit{smooth approximation} by using an \textit{exponential barrier function} with a tunable parameter $\lambda^\text{V}$ as: 
\begin{equation}
\begin{aligned}
c_{i}^{\text{V},(t)} \approx \exp({-\lambda^{\text{V}} (v^{\mathrm{Max}} - \|v_i^{(t)}\|)}).
\end{aligned}
\label{eq:cvti_approx}
\end{equation}
% \vspace{-1em}
\noindent\subsubsection{Collision Penalty with MA-FRS}
To ensure safety under uncertainty, we penalize FRS overlaps via \( c_{i,j}^{\text{FRS},(t)} \) and \( c_{i,j}^{\text{FRS},(T)} \) in \eqref{eq:LiLftilde1}–\eqref{eq:LiLftilde2}. In particular, the terminal penalty \( c_{i,j}^{\text{FRS},(T)} \) encourages well-separated final states, allowing agents to safely stop near their goals despite disturbances. We next derive the MA-FRS model and the collision penalty term $c_{i,j}^{\text{FRS},(t)}$.

\textbf{\textit{(a) MA-FRS Derivation:}} For agent $i\in\mathcal{N}$, let its nominal state, nominal control input, disturbance and error state at time step $t$ be denoted by $\overline{x}_i^{(t)}$, $\overline{u}_i^{(t)}$, ${w}_i^{(t)}\in\mathbb{R}^{n_w}$, and ${e}_i^{(t)}:={x}_i^{(t)}-\overline{x}_i^{(t)}$, where $n_w$ denotes the number of the channels (i.e., independent disturbance components acting along a specific dimension or input direction in the agent's dynamics) of ${w}_i^{(t)}$. We consider the dynamics around $\overline{x}_i^{(t)}$ and $\overline{u}_i^{(t)}$, approximated via first-order linearization as:
\begin{equation}
\begin{aligned}
\dot{x}_i^{(t)} &= \tilde{f}_i\left({x}_i^{(t)}, {u}_i^{(t)}, {w}_i^{(t)}\right)\\
&\approx A_i^{(t)}{x}_i^{(t)}+B_i^{(t)}{u}_i^{(t)}+D_i^{(t)}{w}_i^{(t)},
\label{eq:sys_linear_model}
\end{aligned}
\end{equation}
\noindent where $A_i^{(t)}$, $B_i^{(t)}$, and $D_i^{(t)}$ are the Jacobian matrices obtained by linearizing the nonlinear dynamics $\tilde{f}_i$ around $(\overline{x}_i^{(t)}, \overline{u}_i^{(t)}, 0)$. We assume a bounded disturbance envelope $\|w_i^{(t)}\| \le \bar{w}$, within which the linearization around the nominal trajectory remains valid. The remainder term from the Taylor expansion is assumed to be negligible in this neighborhood.
\begin{figure}
% \vspace{-6mm}
    \centering
    \includegraphics[width=0.9\linewidth]{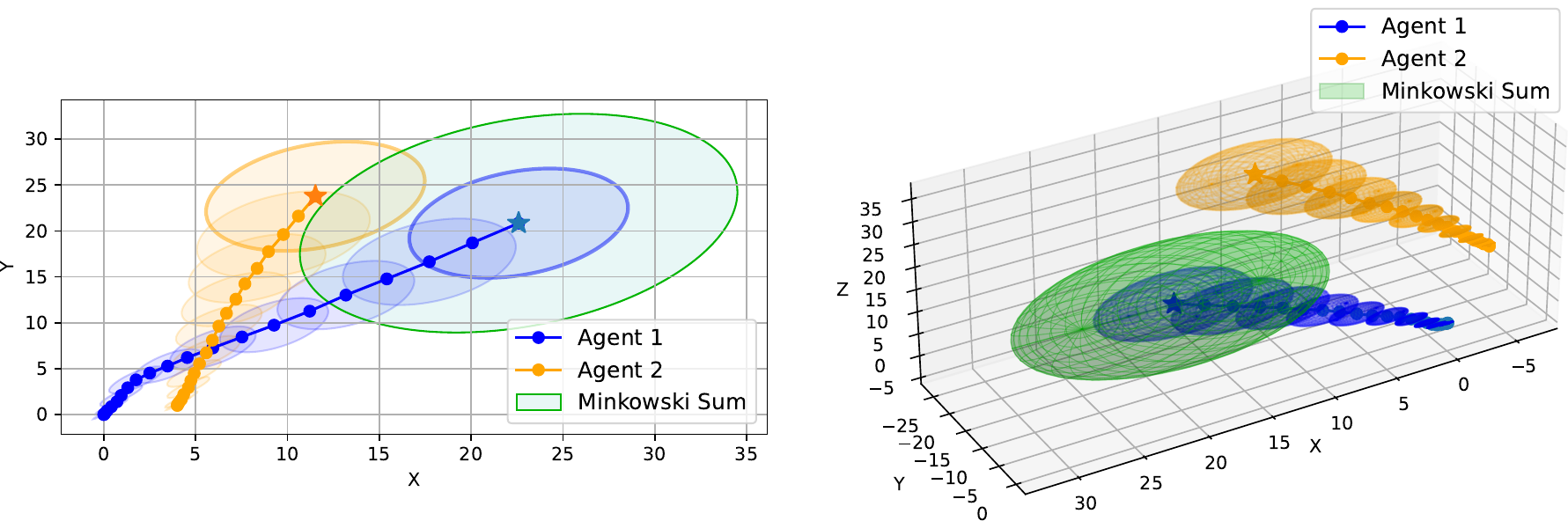}

    \caption{Examples of 2D and 3D ellipsoidal FRSs along the trajectories of two double integrator agents. The green ellipsoids represent the Minkowski sum of the agents’ FRSs at the end of their trajectories.
    % \vspace{-1mm} 
    }
    \label{fig:frs_show}
\end{figure}
Let $K_i^{(t)}$ denote the feedback control gain, where the control policy is  $u^{(t)}_i=K_i^{(t)}{e}_i^{(t)}+\overline{u}_i^{(t)}$. 
To compute the feedback control gain \( K_i^{(t)} \), we use the Linear Quadratic Regulator (LQR), a standard method in optimal control based on the Riccati equation \cite{safonov2003gain}. This provides a reasonable and tractable implementation to obtain the feedback gain. Also, we apply an \textit{ellipsoidal approximation} for the error state FRS, which simplifies the cost calculation by exploiting the tractable intersection condition of ellipsoid geometry. Specifically, the initial error state FRS is denoted as an ellipsoidal set \( \mathcal{E}_i^{(0)} \), centered at the initial state with a positive definite shape matrix \( Q^{(0)}_i \), following Definition \ref{def:ellipsoid}. 

To account for external disturbances, we define \( \tilde{Q}^{(t)}_i \) as the shape matrix for the disturbance-induced error state FRS. The matrix \( \tilde{Q}^{(t)}_i \) is computed by first solving the Lyapunov equation for \( \tilde{Q}^{(t)}_{i,m_w} \), which characterizes the \( m_w \)-th disturbance channel:
\begin{equation} \begin{aligned} -\Phi^{(t)}_i \left(\tilde{Q}^{(t)}_{i,m_w} - \eta t^2 I \right) - \left( \tilde{Q}^{(t)}_{i,m_w} - \eta t^2 I \right) {\left(\Phi^{(t)}_i\right)}^\mathsf{T} \\ = \exp(-t \Phi^{(t)}_i) N_{i,m_w}^{(t)} \exp(-{t \Phi^{(t)}_i}^\mathsf{T}) - N_{i,m_w}^{(t)}. \label{eq:Qittilde} \end{aligned} \end{equation}
The total disturbance-induced FRS shape matrix \( \tilde{Q}^{(t)}_i \) is then approximated by the Minkowski sum of individual disturbance channels: \( \tilde{Q}^{(t)}_i = \mathop{\boxplus}\limits_{m_w=1}^{n_w} \hat{Q}^{(t)}_{i,m_w} \). The total FRS shape matrix \( {Q}^{(t)}_i \) is obtained by combining and propagating these components through the system dynamics:
\begin{equation}
{Q}^{(t)}_{i}= 
\exp{\left(t\Phi^{(t)}_i\right)}\left(Q_i^{(0)} \boxplus \tilde{Q}^{(t)}_{i}\right)\exp \left({t\Phi^{(t)}_i}^\mathsf{T} \right).
\label{eq:Qit}
\end{equation}

\noindent By applying \eqref{eq:Qittilde}–\eqref{eq:Qit} over the planning horizon, we compute the FRS sequence $\{ \mathcal{E}_i^{(t)} \}_{t=0}^T$ for each agent $i$. 
Each agent's FRS shape $\tilde{Q}_i^{(t)}$ is computed via small ($n_p{\times}n_p$) matrix operations, with $n_p$ denoting the position dimension.
The total MA-FRS at each time step $t$ is then defined as the union $\mathcal{E}^{(t)} = \bigcup_{i \in \mathcal{N}} \boldsymbol{\mathcal{E}}_i^{(t)}$ (i.e., the collective reachable space of all agents).

\textbf{\textit{(b) Collision Penalty Calculation:} }
Before designing the collision penalty, we first introduce the following theorem:

\begin{theorem}[Ellipsoidal intersection criterion]
\label{theorem:ellipsoid_intersect}
Let \( \mathcal{E}_1,\mathcal{E}_2 \) be two ellipsoids centered at $c_1, c_2$ with shape matrices $Q_1,Q_2$. The two ellipsoids intersect if and only if $c_2 - c_1 \in \mathcal{T}_O(\mathcal{E}_1) \oplus \mathcal{T}_O(\mathcal{E}_2)$, where $O$ represents the origin.
\end{theorem}

\noindent\textit{Proof: }We can translate \( \mathcal{E}_1 \) and \( \mathcal{E}_2 \) and center them at $O$ as:
\begin{equation}
\hspace{-2mm}
\begin{aligned}
\mathcal{T}_O(\mathcal{E}_1) = \{ x \mid x^\top Q_1 x \leq 1 \}, \quad \mathcal{T}_O(\mathcal{E}_2) = \{ y \mid y^\top Q_2 y \leq 1 \}.
\label{eqs:trans_E1E2}
\end{aligned}
\hspace{-2mm}
\end{equation}
\noindent If $c_2 - c_1 \in \mathcal{T}_O(\mathcal{E}_1) \oplus \mathcal{T}_O(\mathcal{E}_2)$, then we have $u\in\mathcal{T}_O({\mathcal{E}_1})$, $-v\in\mathcal{T}_O({\mathcal{E}_2})$ (equivalent to $v\in\mathcal{T}_O({\mathcal{E}_2})$ due to the symmetry property of $\mathcal{T}_O({\mathcal{E}_2})$ with respect to the origin), and $c_2-c_1=u+(-v)$. Then, we let $z'=u+c_1=v+c_2$, and obtain:
\begin{equation}
\begin{aligned}
u=z'-c_1, v=z'-c_2.
\label{eqs:substi}
\end{aligned}
\end{equation}
We use \eqref{eqs:substi} in \eqref{eqs:trans_E1E2} and get $(z' - c_1)^\top Q_1 (z' - c_1) \leq 1, ~~ (z' - c_2)^\top Q_2 (z' - c_2) \leq 1$, meaning that there exists a point $z'\in\mathbb{R}^n$ such that $z'\in\mathcal{E}_1$ and $z'\in\mathcal{E}_2$. Thus, $\mathcal{E}_1$ and $\mathcal{E}_1$ intersect.

Conversely, if $\mathcal{E}_1$ and $\mathcal{E}_2$ intersect, there exists \( z \in \mathbb{R}^n \) and:
\begin{equation}
\begin{aligned}
(z - c_1)^\top Q_1 (z - c_1) \leq 1 \quad \text{and} \quad (z - c_2)^\top Q_2 (z - c_2) \leq 1.
\end{aligned}
\end{equation}
Let \( p = z - c_1\) and \( q = z - c_2\), we have $c_2 - c_1 = p + (-q)$. 
From \eqref{eqs:trans_E1E2}, we know that $p\in\mathcal{T}_O(\mathcal{E}_1)$ and $q\in\mathcal{T}_O(\mathcal{E}_2)$. Since $\mathcal{T}_O(\mathcal{E}_2)$ is symmetric with respect to the origin $O$, $q\in\mathcal{T}_O(\mathcal{E}_2)$ is equivalent to $-q\in\mathcal{T}_O(\mathcal{E}_2)$. We let $r=-q$, then the condition for intersection can be rewritten as:
\begin{equation}
\begin{aligned}
c_2 - c_1 = p + r, \quad p \in \mathcal{T}_O(\mathcal{E}_1), \quad r \in \mathcal{T}_O(\mathcal{E}_2).
\end{aligned}
\end{equation}
\noindent Combining Definition \ref{def:Minkowski_Sum}, we get
\begin{equation}
\begin{aligned}
c_2 - c_1 \in \mathcal{T}_O(\mathcal{E}_1) \oplus \mathcal{T}_O(\mathcal{E}_2). 
\label{eq:cond1}
\end{aligned}
\end{equation}
Thus, we have proven that the two ellipsoids \( \mathcal{E}_1 \) and \( \mathcal{E}_2 \) intersect if and only if \( c_2 - c_1 \in \mathcal{T}_O(\mathcal{E}_1) \oplus \mathcal{T}_O(\mathcal{E}_2) \). $\hfill\blacksquare$

\begin{remark}[Translation simplification]
To simplify the computations, for the condition in \eqref{eq:cond1}, we can translate $\mathcal{T}_O(\mathcal{E}_1) \oplus \mathcal{T}_O(\mathcal{E}_2)$ to be centered on $c_1$, yielding:
$c_2\in\mathcal{T}_{c_1}(\mathcal{T}_O(\mathcal{E}_1) \oplus \mathcal{T}_O(\mathcal{E}_2))$. Fig.~\ref{fig:frs_show} illustrates this geometric relationship, where $c_1$ and $c_2$ correspond to the positions of agent 1 and 2, and the green ellipsoid centered at $c_1$ corresponds to $\mathcal{T}_{c_1}(\mathcal{T}_O(\mathcal{E}_1) \oplus \mathcal{T}_O(\mathcal{E}_2))$. Then, the condition that agent 2 is outside the green ellipsoid is equivalent to that the ellipsoidal FRSs of the agents do not intersect.
\end{remark}

% \begin{figure}[!t]
% \vspace{-1mm}
% \begin{algorithm}[H]
% \caption{Solution for finding the $\varepsilon$-NE of our DPG}\label{alg:dpg_solution}
% \KwIn{Initial strategy $\mathbf{u}^0\gets(\mathbf{u}_1^0, \mathbf{u}_2^0, \ldots, \mathbf{u}_N^0)$, $k\gets 0$, convergence threshold $\varepsilon$, max iteration number $\overline{k}$}
% \KwOut{Nominal control sequence $\boldsymbol{\mathbf{u}}^{k+1}$}

% \hspace{0.5em}Calculate the FRSs of all agents based on \eqref{eq:Qittilde}-\eqref{eq:Qit}\\
% \hspace{1.0em}Compute initial accumulated PF value based on $\mathbf{u}^0$:
% \[
% \mathcal{J}^0 = \sum_{t=0}^{T-1} P(\boldsymbol{x}^{(t)}, \boldsymbol{u}^{(t)}) + R(\boldsymbol{x}^{(T)})
% \]

% ~\While{$\left| \mathcal{J}^{k} - \mathcal{J}^{k-1} \right| \geq \varepsilon$ \textbf{or} $k\leq \overline{k}$}{
%     ~\For{$i \gets 1$ \textbf{to} $N$}{
%         ~Solve ${\boldsymbol{\tilde{\mathcal{P}}}}^k_i$ (with $\tilde{J_i}(\cdot,\cdot)$ in \eqref{eq:tilde_Ji} and its components detailed in \eqref{eq:trackingcost1}\eqref{eq:trackingcost2}\eqref{eq:cvti_approx}\eqref{eq:cfrs_f}) by MPC to obtain $\mathbf{u}_{i}^*$\;
%     }
%     ~$\boldsymbol{\mathbf{u}}^{k+1} \gets (\mathbf{u}_{i}^*, {\mathbf{u}}_{-i}^{k})$\;\\
%     ~Compute $\mathcal{J}^{k}$ (with the same formula as $\mathcal{J}^0$)\;\\
%     ~$k \gets k + 1$\;
% }
% \end{algorithm}
% \end{figure}

% \begin{figure}[!t]
% \vspace{-1mm}
\begin{algorithm}
\caption{Solution for computing an $\varepsilon$-NE of our DPG}\label{alg:dpg_solution}
\KwIn{Initial profile $\mathbf{u}^0=(\mathbf{u}_1^0,\ldots,\mathbf{u}_N^0)$, tolerance $\varepsilon>0$, max iterations $\overline{k}$}
\KwOut{Nominal control sequence $\mathbf{u}^*$}

\hspace{0.5em}Compute FRS for all agents using \eqref{eq:Qittilde}--\eqref{eq:Qit}\;, $k\!\gets\!0$\;

\While{$k<\overline{k}$}{
  \For{$i \gets 1$ \KwTo $N$}{
    Solve ${\boldsymbol{\tilde{\mathcal{P}}}}^k_i$ (use $\tilde{J}_i$ in \eqref{eq:tilde_Ji}, with components \eqref{eq:trackingcost1}, \eqref{eq:trackingcost2}, \eqref{eq:cvti_approx}, \eqref{eq:cfrs_f}) by MPC to obtain a candidate $\mathbf{u}_{i}^{\star}$\;
    
    Compute improvement
    \[
      r_i^k \;\gets\; J_i\!\big(\boldsymbol{x}^{(0)},(\mathbf{u}_i^{k},\mathbf{u}_{-i}^{k})\big)
      - J_i\!\big(\boldsymbol{x}^{(0)},(\mathbf{u}_i^{\star},\mathbf{u}_{-i}^{k})\big)
    \]
  }
  \eIf{$\max_{i\in\mathcal N} r_i^k < \varepsilon$}{
    $\mathbf{u}^* \gets \mathbf{u}^k$; \textbf{break}\;
  }{
    Select any $i_k \in \arg\max_{i} r_i^k$,
    $\mathbf{u}^{k+1} \gets (\mathbf{u}_{i_k}^{\star},\,\mathbf{u}_{-i_k}^{k})$\;
    
    $k \gets k+1$\;
  }
}
\end{algorithm}
% \end{figure}

We now proceed to design the collision penalty for safe motion planning, focusing on maintaining separation between agents' FRSs to prevent collisions. We consider only position-related uncertainties which causes primary collision risks. For each agent \( i \), we extract the position vector \( p_i^{(t)} \) from \( x_i^{(t)} \) and define \( \hat{Q}^{(t)}_{i,\text{pos}} \) as the submatrix of \( Q^{(t)}_i \) corresponding to position dimensions. Therefore, for neighboring agents $i$ and $j$, the resulting FRS ellipsoids \( \mathcal{E}_{i,\text{pos}}^{(t)} \) and \( \mathcal{E}_{j,\text{pos}}^{(t)} \) must not intersect at each time step. Based on Theorem \ref{theorem:ellipsoid_intersect} and Definition \ref{def:ellipsoid}, the following collision avoidance condition can then be derived:
\begin{equation}
\hspace{-2.85mm}
\begin{aligned}
\xi^{(t)}_{i,j}=(p_i^{(t)}-p_j^{(t)})^\mathsf{T}(\hat{Q}^{(t)}_{i,\text{pos}}\boxplus \hat{Q}^{(t)}_{j,\text{pos}})^{-1}(p_i^{(t)}-p_j^{(t)})-1>0, 
\end{aligned}
\label{xiij}
\hspace{-2mm}
\end{equation}
where $\hat{Q}^{(t)}_{i,\text{pos}} \boxplus \hat{Q}^{(t)}_{j,\text{pos}}$ is the shape matrix of $\mathcal{T}_O(\mathcal{E}_{i,\text{pos}}^{(t)})\oplus\mathcal{T}_O(\mathcal{E}_{j,\text{pos}}^{(t)})$.
Similar to \eqref{eq:cvti}, we encode this condition using a barrier function, defining the collision penalty $c_{i}^{\text{FRS},(t)}$ as:  
\begin{equation}
c_{i,j}^{\text{FRS},(t)} \approx \exp({-\lambda^{\text{FRS}} \xi^{(t)}_{i,j}}) ,
\label{eq:cfrs_f}
\end{equation}
\noindent where $\lambda^{\text{FRS}}$ is tunable. These formulations allow us to embed a differentiable, uncertainty-aware collision penalty into the DPG cost structure, preserving tractability while ensuring safety. Alternatively, neighbors can be defined anisotropically using the MA-FRS shape matrices, i.e., $\tilde{\mathcal{N}}^{(t)}_i = \{j~|~(p_i^{(t)}\!-\!p_j^{(t)})^\top (Q_i^{(t)}\boxplus Q_j^{(t)})^{-1}(p_i^{(t)}\!-\!p_j^{(t)})\le 1,\; j\neq i\}$, consistent with the ellipsoidal collision metric in \eqref{eq:cfrs_f}. This provides a conservative and uncertainty-aware alternative to isotropic Euclidean screening.

With the enhanced cost functions, tracking and velocity terms depend only on each agent’s local state and input, while the MA-FRS–based collision term introduces symmetric neighbor coupling. Thus, the resulting costs satisfy the structural conditions of Definition~\ref{def:DPG}, preserving the DPG structure for scalable motion planning under uncertainty.
% With the enhanced cost functions defined, we observe that the tracking and velocity penalties depend only on each agent’s local state and input, while the MA-FRS–based collision term $c_{i,j}^{\text{FRS},(t)}$ introduces symmetric coupling between neighbors. As such, the resulting costs $\tilde{L}_i(\cdot)$ and $\tilde{L}_i^\text{F}(\cdot)$ satisfy the structural conditions of Definition~\ref{def:DPG}, thereby preserving the DPG formulation for scalable motion planning under uncertainty.

% With the composition of the enhanced cost functions established, it is clear that the tracking cost terms \eqref{eq:trackingcost1}-\eqref{eq:trackingcost2} and velocity constraint penalties \eqref{eq:cvti_approx} depend solely on each agent’s own state and control input. In contrast, the collision penalty based on MA-FRS,
% $c_{i,j}^{\text{FRS},(t)}$, serves as the coupling cost between neighboring agents. As shown in \eqref{xiij}–\eqref{eq:cfrs_f}, $c_{i,j}^{\text{FRS},(t)}$ satisfies the symmetry assumption imposed in \eqref{Li_c}–\eqref{Lif_c}. In summary, our proposed enhanced stage and terminal costs $\tilde{L}_i(\cdot)$ and $\tilde{L}_i^\text{F}(\cdot)$ fully adhere to the structural conditions specified in Definition \ref{def:DPG}. Consequently, the proposed cost formulation preserves the integrity of the DPG framework, thereby enabling scalable multi-agent motion planning under uncertainty.
% \vspace{-1em}
\subsection{Reaching the $\varepsilon$-NE}\label{Sec:NE}
This section operationalizes our framework by detailing how we iteratively compute a safe and robust $\varepsilon$-NE (i.e., the nominal control sequence for all agents) using ND-iBR within the DPG framework (the inclusion of MA-FRS penalties dynamically adjusts agent strategies, ensuring convergence to the $\varepsilon$-NE while respecting safety constraints under disturbances). Specifically, the MA-FRS penalties enforce non-overlapping safety regions, significantly improving the convergence behavior and guiding agents toward stable, collision-free solutions. Our procedure to obtain the $\varepsilon$-NE is outlined in Algorithm \ref{alg:dpg_solution} and detailed below.

\begin{figure*}
% \vspace{-6.25mm}
        \begin{subfigure}
        \centering
        \includegraphics[width=1.0\linewidth]{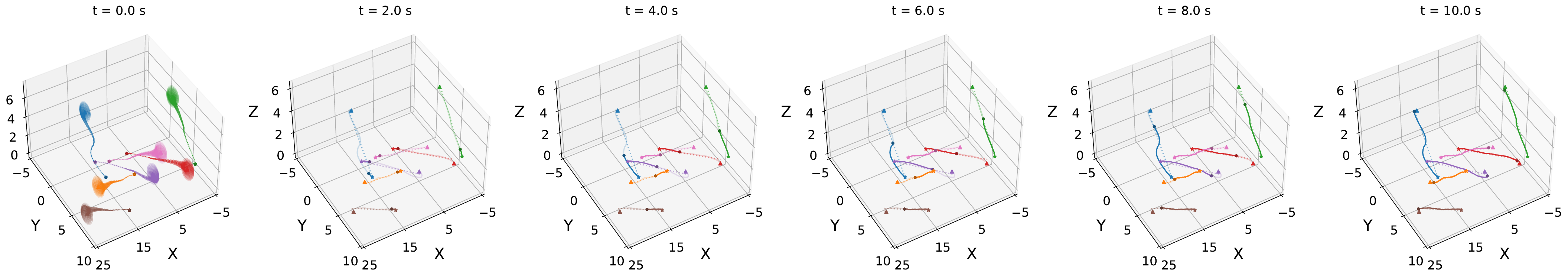}
        \caption{Example trajectories of 7 quadrotors with random start positions and goal positions (denoted by star and triangle marks, respectively). Dashed lines represent the reference trajectories and solid lines represent the final trajectories under truncated Gaussian noise with variance $0.02$. An illustration of MA-FRSs along the trajectory is shown at the initial time step ($t$ = 0).}
        \label{fig:agent7}
        \end{subfigure}
        \begin{subfigure}
                \centering
    \includegraphics[width=1.0\linewidth]{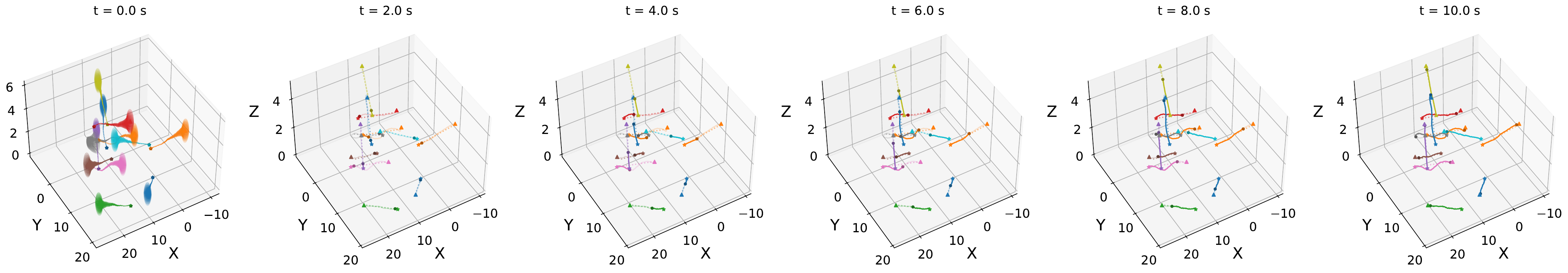}
      % \vspace{-6mm}
\caption{Example trajectories of 12 quadrotors under the same settings as the scenario shown in Fig.~\ref{fig:agent7}. This denser configuration induces more frequent neighborhood changes, as agents enter and exit each other's proximity range throughout execution. The effectiveness of RE-DPG under this setting signals its robustness to dynamic, time-varying neighbor structures.
% \vspace{-6mm}
}
    \label{fig:agent12}
        \end{subfigure}
\end{figure*}

\begin{figure}[htbp]
    \centering
    \begin{subfigure}
        \centering
        \includegraphics[width=0.47\linewidth]{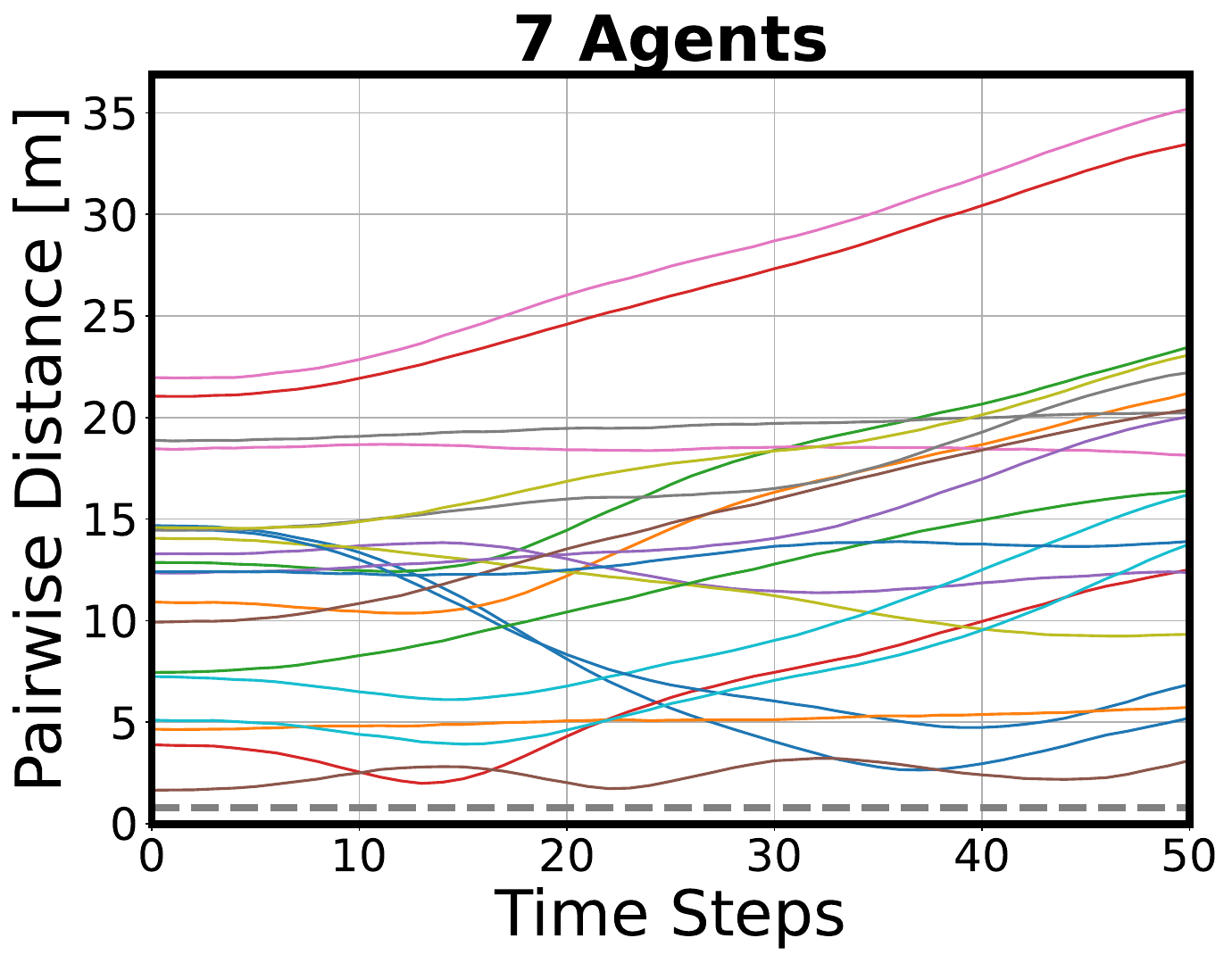}
        \label{fig:sub1}
    \end{subfigure}
    \begin{subfigure}
        \centering
        \includegraphics[width=0.47\linewidth]{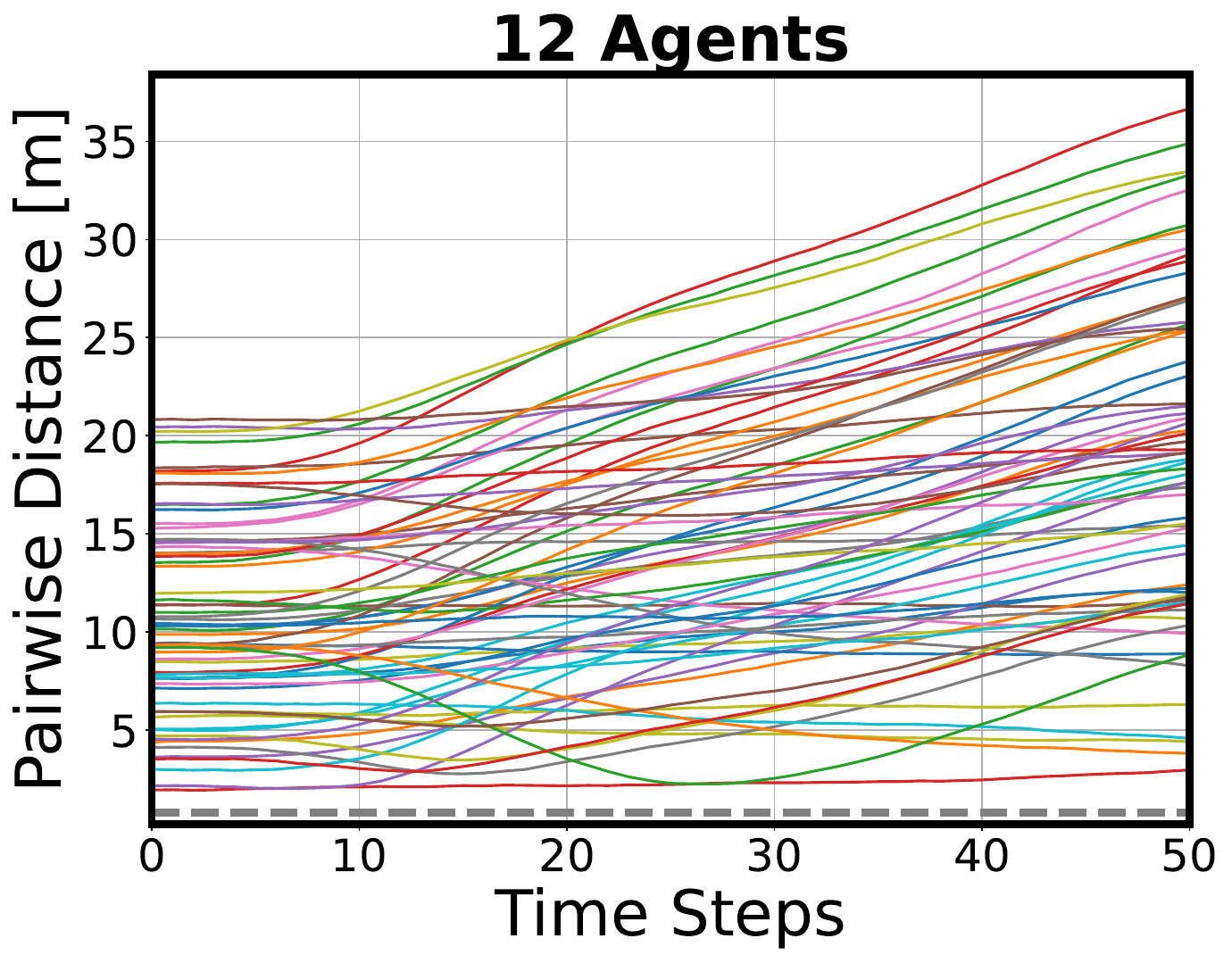}
        \label{fig:sub2}
    \end{subfigure}
% \vspace{-3mm}
    \caption{Pairwise distances between agents regarding the settings of 7 agents and 12 agents related to Fig. \ref{fig:agent7} and Fig. \ref{fig:agent12}, respectively. The dashed line at the bottom of each plot corresponds to the desired distance threshold.}
    \label{fig:pairwise_dist}
\end{figure}

We solve each agent-specific subproblem \( \boldsymbol{\tilde{\mathcal{P}}}^k_i \) using Model Predictive Control (MPC), which applies only the first input of a re-solved finite-horizon plan at each step. While open-loop NE solutions offer computational efficiency but lack robustness, and closed-loop strategies offer robustness at higher cost, MPC strikes a balance by approximating closed-loop behavior in a tractable way. This integrated formulation combines the strengths of game-theoretic planning and real-time MPC, improving both robustness and safety guarantees \cite{bhatt2023efficient}. It also enables online correction for accumulated errors or unmodeled disturbances through continual state feedback. For efficient optimization, we use Interior Point OPTimizer (IPOPT) \cite{wachter2006implementation}.

% To solve each agent-specific optimal control problem \( \boldsymbol{\tilde{\mathcal{P}}}^k_i \) within the ND-iBR framework, we employ Model Predictive Control (MPC), which computes collision-free control inputs by solving a finite-horizon optimal control problem using a receding horizon methodology. For efficient optimization, we use the IPOPT solver \cite{wachter2006implementation}, which is well-suited for large-scale nonlinear programming in multi-agent settings. Here, we take advantage of an additional advantage of integrating MPC when considering the nature of the NE strategy. As introduced in Definition \ref{def:open_loop_NE}, an open-loop NE offers computational efficiency but lacks robustness due to the absence of state feedback. In contrast, a closed-loop strategy continuously updates control inputs based on real-time state information, which enhances robustness but increases computational demand. The receding horizon structure of MPC effectively approximates a closed-loop strategy, offering a balance between computational tractability and real-time adaptability. This integrated solution not only improves robustness but also ensures safety guarantees, marking a notable advancement over the common isolated applications of MPC or game-theoretic planning \cite{bhatt2023efficient}.
\begin{figure*}
% \vspace{-6.8mm}
\centering
\includegraphics[width=0.95\linewidth]{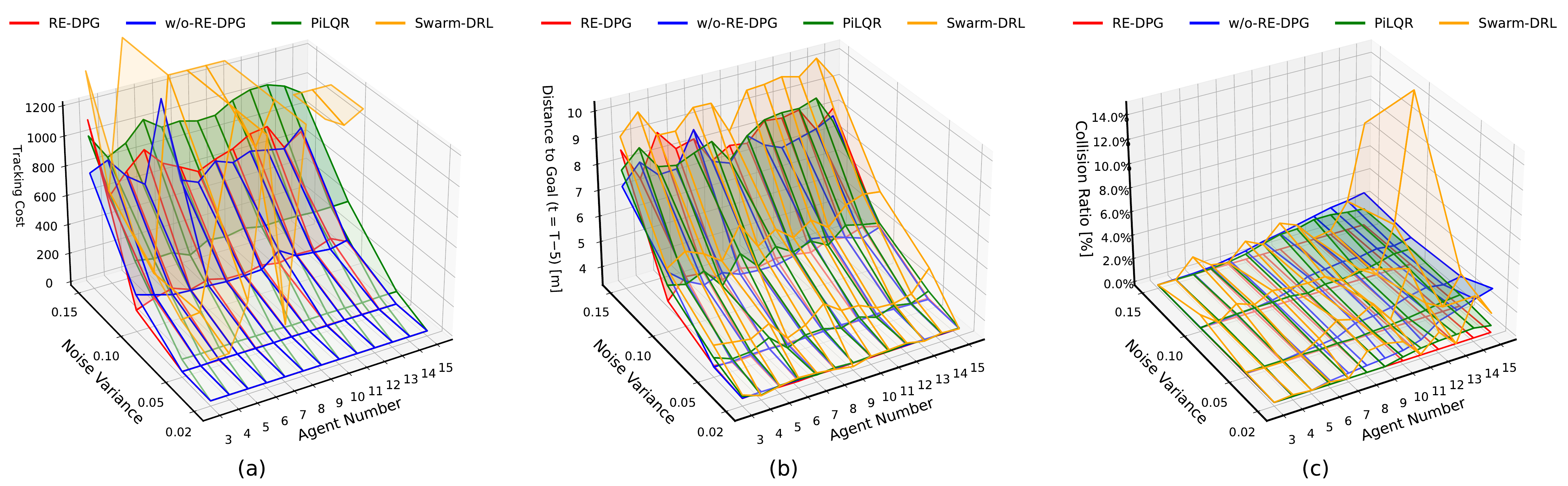}
    \caption{Comparison results across three evaluation metrics with varying numbers of agents and disturbance levels.}
    \label{fig:allmetrics}
\end{figure*}
% \vspace{-0.5em}
\section{Numerical Simulations}\label{sec:simulation}

In this section, we perform numerical simulations to validate the effectiveness of our RE-DPG methodology. We consider a 3D environment of size $30\,\text{m} \times 30\,\text{m} \times 10\,\text{m}$, where multiple quadrotor agents (see Appendix for dynamics) navigate from their initial to target positions. The total planning horizon spans 50 time steps with a time interval of 0.2s, and nominal trajectories are computed using an MPC scheme with a 20-step receding horizon. We perform Monte Carlo simulations by varying the number of agents \( N \in [3, 15] \) and the variance \( \sigma \in \{0.02, 0.05, 0.10, 0.15\} \) of the Gaussian noise in each agent's dynamics (i.e., disturbance in \eqref{eq:sys_linear_model}). The noise is zero-mean and truncated within the range \( [-\sigma, \sigma] \) to reflect bounded conditions. For each configuration, 50 independent runs are performed, and results are averaged for statistical reliability. Agents' initial positions and destinations are randomly assigned with a minimum separation of $1.0\, \text{m}$ to avoid artificial collisions at initialization, and the maximum velocity is set to \( v^{\mathrm{Max}} = 5\,\text{m/s} \). 
We use Euclidean distance in Section~\ref{Sec:iBR} to define neighbor sets in all simulations for efficiency and select $\varepsilon=10^{-2}$ in Algorithm~\ref{alg:dpg_solution}. To conduct performance comparisons, we consider the following baselines:

\noindent\textbullet~\textit{w/o-RE-DPG}: This variant removes the MA-FRS penalty term \eqref{eq:cfrs_f} and replaces it with a Euclidean proximity-based penalty activated when \(d < d^{\mathrm{Col}}\), serving as an ablation to isolate the contribution of MA-FRS shaping.

\noindent\textbullet~\textit{Probabilistic iLQR (PiLQR)}: A probabilistic-based robust motion planning method \cite{lembono2021probabilistic} is considered to solve \( \boldsymbol{\tilde{{\mathcal{P}}}^k_i} \) without the MA-FRS penalty term. 

\noindent\textbullet~\textit{Swarm-DRL}: An end-to-end, {decentralized} deep reinforcement learning (DRL) method for swarms introduced in \cite{huang2024collision}.

% To evaluate the robustness of different methods under external disturbances, we employ an MPC controller to track the nominal trajectories generated by each approach. We first visualize the resulting agent trajectories in 3D space for two scenarios: one with 7 agents (Fig. \ref{fig:agent7}) and another with 12 agents (Fig. \ref{fig:agent12}). These visualizations show that all agents safely reach their destinations, demonstrating the effectiveness of the \textit{RE-DPG} framework. This performance is primarily attributed to the MA-FRS-based penalty term, which helps agents avoid potential collision zones while progressing toward their destinations, thereby mitigating the risk of collisions. 
% For the hyperparameters, we set $\lambda^{\mathrm{FRS}}=\lambda^{\mathrm{V}}{=}10$, $Q_i{=}\mathrm{diag}(\mathbf{0^{1\times 9}}, 10, 10, 10)$, and $R_i{=}\mathrm{diag}(1,1,1,1)$ for all agents across scenarios.
% To further validate the ability of \textit{RE-DPG} to ensure safety and robustness in dynamic and uncertain environments, Fig. \ref{fig:pairwise_dist} shows the pairwise distances between agents for the two previously discussed scenarios, with the gray dashed line indicating the collision threshold \(d^\text{Col} = 0.5\,\text{m}\). As shown in Fig. \ref{fig:pairwise_dist}, all distances stay above this threshold, indicating that no collisions occur throughout the process. To further evaluate our \textit{RE-DPG}, we study the following three key performance metrics:

To evaluate robustness under external disturbances, we apply an MPC controller to track the nominal trajectories generated by each method. Figs.~\ref{fig:agent7} and \ref{fig:agent12} show example trajectories for 7 and 12 agents, respectively, where all agents safely reach their targets. In terms of runtime, the MA-FRS propagation requires on average 102\,ms per agent and is computed only once during a trial. Under homogeneous disturbance assumptions, shape computations can be shared across agents, keeping the aggregate cost nearly constant as $N$ increases.
This highlights the effectiveness of our \textit{RE-DPG} framework, largely due to the MA-FRS penalty that promotes collision-free progress toward goals.
We set $\lambda^{\mathrm{FRS}}=\lambda^{\mathrm{V}}=10$, $Q_i = \mathrm{diag}(\mathbf{0^{1\times 9}},10,10,10)$, and $R_i = \mathrm{diag}(1,1,1,1)$ for all agents, where $\mathrm{diag}(\cdot)$ denotes a diagonal matrix with the specified entries on its diagonal. Fig.~\ref{fig:pairwise_dist} confirms safety by showing that all pairwise distances remain above the collision threshold $d^{\mathrm{Col}} = 0.5$ m. We further assess RE-DPG through three performance metrics:

% \noindent \textbf{\textit{1) Performance Evaluation on Tracking Cost:}} 
% The tracking cost measures how well agents follow their planned trajectories, reflecting efficiency and control accuracy. It is computed based on \eqref{eq:trackingcost1}-\eqref{eq:trackingcost2}, with identity weighting matrices \( \boldsymbol{Q}_i, \boldsymbol{Q}^\text{F}_i \), and \( \boldsymbol{R}_i \) for unbiased comparison. As shown in Fig.~\ref{fig:allmetrics}(a), all methods perform well under low noise. However, as the noise variance increases, the tracking cost for \textit{PiLQR} rises significantly, especially in scenarios with more agents. In contrast, both \textit{RE-DPG} and its variant \textit{w/o-RE-DPG} maintain stable performance across various noise levels. While \textit{RE-DPG} incurs a slightly higher cost due to the MA-FRS penalty term, this additional overhead reflects proactive safety adjustments. Overall, \textit{RE-DPG} achieves a good balance between robustness, safety, and efficient trajectory tracking in dynamic environments.

\noindent \textbf{\textit{1) Performance Evaluation on Tracking Cost:}} 
The tracking cost measures deviation from a nominal reference and is computed from \eqref{eq:trackingcost1}–\eqref{eq:trackingcost2} with identity \(Q_i,Q_i^{\mathrm F},R_i\) for an unbiased comparison. 
As shown in Fig.~\ref{fig:allmetrics}(a), \textit{PiLQR}'s cost grows markedly with noise and team size, whereas \textit{RE-DPG}'s and \textit{w/o-RE-DPG}'s costs remain stable; the slight extra cost of \textit{RE-DPG} simply reflects MA-FRS safety shaping. \textit{Swarm-DRL} exhibits consistently higher cost values, as it does not explicitly follow a predefined reference trajectory. Instead, its reward function primarily prioritizes collision avoidance, while the low-level thrust commands it generates may lead to oscillatory or circuitous movements, thereby inflating the quadratic error metric as in \eqref{eq:trackingcost1} and \eqref{eq:trackingcost2}. To sum up, our \textit{RE-DPG} remains stable and competitive across noise levels and team sizes.

\noindent \textbf{\textit{2) Performance Evaluation on Time Efficiency:}}  
Time efficiency is measured by the remaining distance to the goal at \(t = T - 5\). 
Fig.~\ref{fig:allmetrics}(b) shows that \textit{w/o-RE-DPG} reaches the goal slightly faster with \textit{RE-DPG} and \textit{PiLQR} closely behind. It replaces uncertainty-aware MA-FRS shaping with a Euclidean penalty, which focuses more on tracking but has reduced safety under disturbances due to looser pairwise margins.
\textit{Swarm-DRL} attains competitive efficiency at low noise; under higher noise or larger teams, its efficiency degrades modestly. Crucially, our \textit{RE-DPG} maintains competitive efficiency while ensuring safety under uncertainty (Fig.~\ref{fig:pairwise_dist}), showing that MA-FRS shaping preserves progress toward goals without sacrificing robustness.

% \noindent \textbf{\textit{3) Performance Evaluation on Safety:}} To assess safety, we compute the average collision rate, defined as the proportion of time steps during which the distance between any pair of agents falls below the threshold \( d^{\text{Col}} \), where a lower average collision rate indicates a  better inter-agent avoidance collision performance. Fig.~\ref{fig:allmetrics}(c) shows that our \textit{RE-DPG} consistently achieves collision-free performance across varying numbers of agents and noise levels as it  integrates MA-FRS to explicitly account for the propagation of state uncertainty, thereby establishing robust safety boundaries. Although \textit{PiLQR} incorporates collision penalties in its cost function, it lacks explicit modeling of safety margins under uncertainty. As a result, \textit{PiLQR} is prone to collisions, particularly in dense agent or high-disturbance environments. The \textit{w/o-RE-DPG}, which omits MA-FRS-based penalties, fails to prevent collisions entirely due to the absence of uncertainty-aware safety mechanisms.

\noindent \textbf{\textit{3) Performance Evaluation on Safety:}}  
We assess safety using the average collision ratio, defined as the proportion of time steps when any pairwise distance falls below \(d^{\mathrm{Col}}\). 
Fig.~\ref{fig:allmetrics}(c) highlights that \textit{RE-DPG} consistently achieves collision-free performance across agent counts and noise levels by explicitly modeling uncertainty propagation and enforcing MA-FRS–based separation. 
The \textit{w/o-RE-DPG} ablation replaces the MA-FRS penalty with a simple Euclidean threshold, which discourages close proximity but lacks awareness of uncertainty propagation, leading to collisions under higher noise.
Although \textit{PiLQR} incorporates uncertainty-aware penalties, it lacks explicit margins and thus suffers from higher collision ratios in dense or high-disturbance settings. Similarly, \textit{Swarm-DRL}, which relies on implicitly learned safety behavior rather than explicit uncertainty modeling, shows degraded safety as noise and agent density increase, further underscoring the importance of MA-FRS–based guarantees in \textit{RE-DPG}.

Overall, our proposed \textit{RE-DPG} achieves a good balance between robustness, safety, and efficient trajectory tracking in dynamic environments.

\begin{figure*}[htbp]
    \centering
    \includegraphics[width=0.9\linewidth]{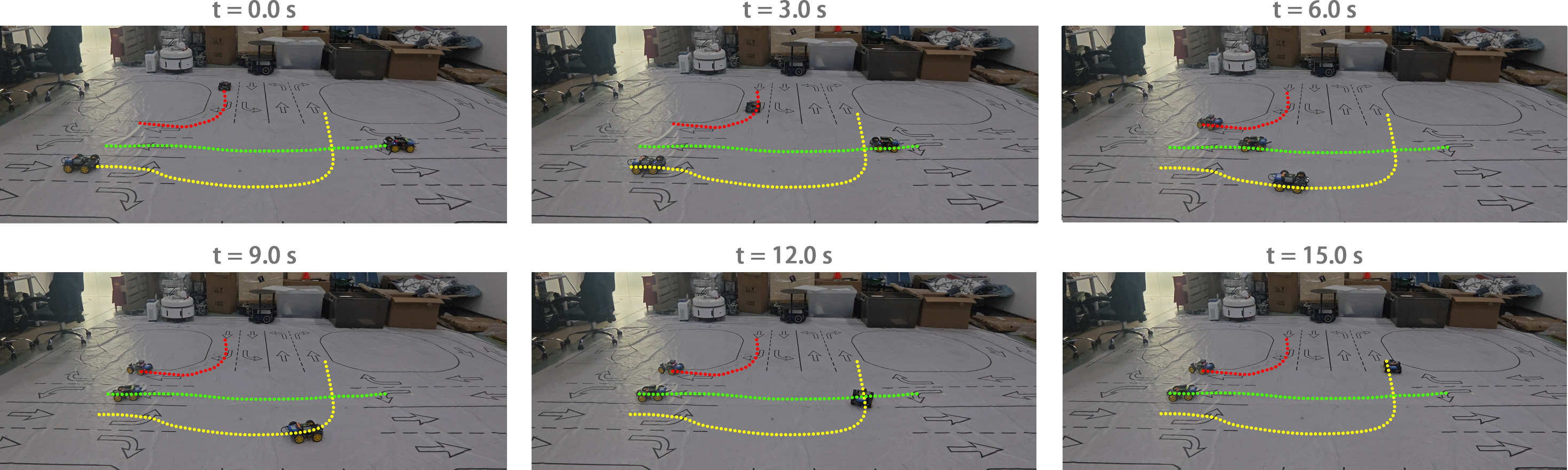}
    \caption{\textit{Scenario 1}: Snapshots and visualization.}
    \label{fig:agent3}
    % \vspace{-2mm}
\end{figure*}

\begin{figure*}[htbp]
    \centering
    \includegraphics[width=0.9\linewidth]{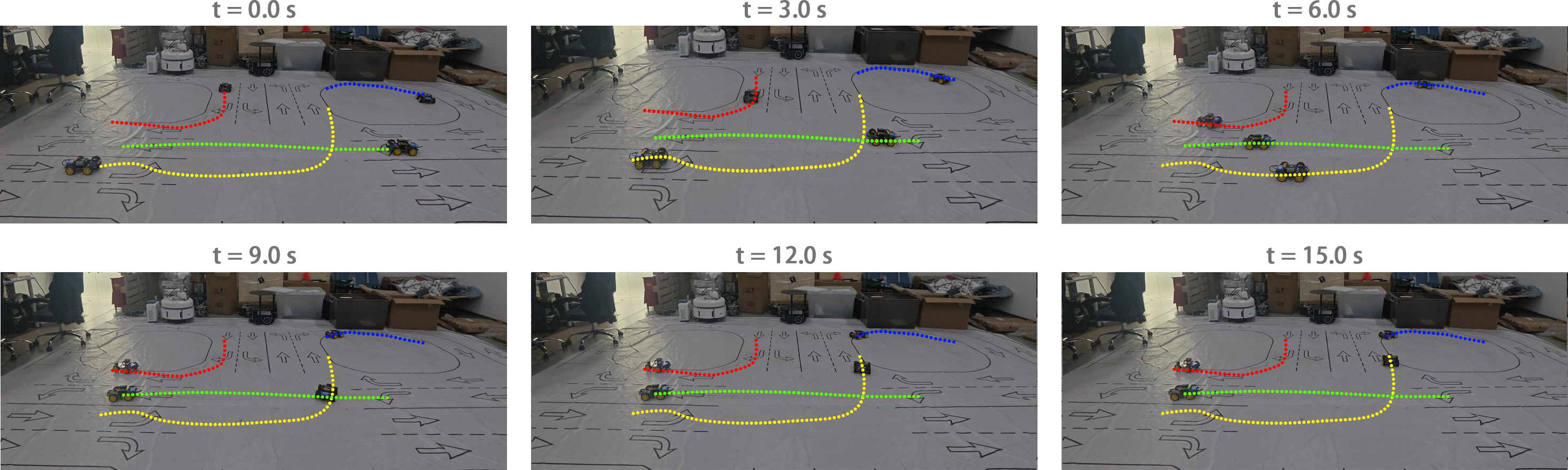}
    \caption{\textit{Scenario 2}: Snapshots and visualization.} 
    \label{fig:agent4}
    % \vspace{-5.5mm}
\end{figure*}

% \vspace{-2mm}
% \section{Real-World Experiments}
% To validate our methodology in real-world settings, we conduct experiments with multiple agents moving across an intersection, simulating 2D urban traffic coordination. The experiments use four-wheel-drive (4WD) vehicles controlled by Raspberry Pi 5, running on the Robot Operating System (ROS) for state sharing and distributed planning. Wheel encoders are used to measure the rotational speed and estimate each vehicle's state in the world coordinates. Two scenarios were designed: \textit{(I) Scenario 1:} Three vehicles interact at an intersection, two of them turning and one going straight (see Fig. \ref{fig:agent3}). \textit{(II) Scenario 2:} A fourth vehicle is involved, but positioned far enough from the others so that its FRS does not overlap with theirs (see Fig. \ref{fig:agent4}). 
% For each scenario, we show the trajectories of vehicles at six time points. As shown in Fig. \ref{fig:agent3}, when the straight-moving vehicle (green) enters the proximity of the turning vehicle (yellow), to avoid a potential collision, the turning vehicle chooses to yield, resulting in an outward trajectory adjustment. Fig. \ref{fig:agent4} shows that the addition of the new vehicle does not affect the original collision-avoidance interaction results. Although the vehicles deviated from their reference trajectories to avoid collisions, they successfully and safely reached their goals.

\section{Real-World Experiments}
To validate our methodology in real-world settings, we conduct experiments with multiple agents moving across an intersection, simulating 2D urban traffic coordination. The experiments use four-wheel-drive (4WD) vehicles controlled by Raspberry Pi~5, running on the Robot Operating System (ROS) for state sharing and distributed planning. Wheel encoders are used to measure the rotational speed and estimate each vehicle's state in the world coordinates. We set the collision threshold to $d^{\mathrm{Col}}=0.3$\,m; each trial lasts $15$\,\text{s} with a $0.2$\,\text{s} control step; and control inputs are generated by an MPC with a $10$-step planning horizon. Other hyperparameters are the same as in Sec.~\ref{sec:simulation}. In addition to the trajectories, we report two quantitative safety metrics: the \emph{minimum pairwise distance} over each trial and the \emph{near-collision count} (number of instants where the pairwise distance is below $d^{\mathrm{Col}}+0.05$\,m). 

Two scenarios were designed: \textit{(I) Scenario 1:} Three vehicles interact at an intersection, two of them turning and one going straight (see Fig.~\ref{fig:agent3}). \textit{(II) Scenario 2:} A fourth vehicle is involved, but positioned far enough from the others so that its FRS does not overlap with theirs (see Fig.~\ref{fig:agent4}). For each scenario, the MA-FRS propagation is calculated only once and requires 46\,ms on average. On-board planning runs in real time (per-step latency $<0.2$\,s) and we show the trajectories of vehicles at six time points. As shown in Fig.~\ref{fig:agent3}, when the straight-moving vehicle (green) enters the proximity of the turning vehicle (yellow), to avoid a potential collision, the turning vehicle chooses to yield, resulting in an outward trajectory adjustment; quantitatively, the minimum pairwise distance is $0.4$\,m ($>d^{\mathrm{Col}}$), the near-collision count is $0$. 
Fig.~\ref{fig:agent4} shows that the addition of the new vehicle does not affect the original collision-avoidance interaction results; the same metrics as in Scenario~1 are observed, with identical values for minimum pairwise distance and near-collision count. Although the vehicles deviated from their reference trajectories to avoid collisions, they successfully and safely reached their goals.

% \vspace{-0.5em}
% \vspace{-2mm}
\section{Conclusion and Future Work}
We introduced a decentralized framework for safe multi-agent motion planning under uncertainty. The problem was formulated as a Dynamic Potential Game (DPG) and solved through a Neighborhood-Dominated iterative Best Response (ND-iBR) scheme, grounded in an iterated $\varepsilon$-best-response ($i\varepsilon$-BR) process to guarantee finite-step convergence to an $\varepsilon$-Nash equilibrium. To ensure safety and robustness, we incorporated Multi-Agent Forward Reachable Sets (MA-FRSs) into the optimization, explicitly modeling uncertainty and enforcing collision-avoidance constraints. Both simulations and hardware experiments validated the effectiveness of our method across dynamic multi-agent scenarios.
Future directions include: \textit{1) Robust planning under partial observability and communication constraints}, where unreliable or intermittent communication may limit neighborhood information; \textit{2) Adaptive robustness against out-of-envelope disturbances}, noting that fixed disturbance bounds may be violated in practice, motivating online disturbance estimation and real-time FRS recomputation to maintain safety under unexpected conditions.

\section*{Appendix -- Dynamic Models}
% \vspace{-0mm}
\textit{1) Quadrotor Model}: At time step \( t\), the quadrotor's state is given by \( x^{(t)} = [\omega_x^{(t)}, \omega_y^{(t)}, \omega_z^{(t)}, \theta_x^{(t)}, \theta_y^{(t)}, \theta_z^{(t)}, v_x^{(t)}, v_y^{(t)}, v_z^{(t)}, p_x^{(t)}, p_y^{(t)}, p_z^{(t)}]^\mathsf{T} \), where \( p_x^{(t)}, p_y^{(t)}, p_z^{(t)} \) are the positions, \( v_x^{(t)}, v_y^{(t)}, v_z^{(t)} \) are the velocities, \( \theta_x^{(t)}, \theta_y^{(t)}, \theta_z^{(t)} \) are the roll, pitch, and yaw angles, and \( \omega_x^{(t)}, \omega_y^{(t)}, \omega_z^{(t)} \) are the corresponding angular velocities. The control input is \( u^{(t)} = [F^{(t)}_{\text{th}}, \tau^{(t)}_x, \tau^{(t)}_y, \tau_z^{(t)}] \), where \( F^{(t)}_{\text{th}} \) is the total thrust, and \( \tau^{(t)}_x, \tau^{(t)}_y, \tau_z^{(t)} \) are the control torques. For the detailed quadrotor dynamic model, we refer to \cite{zhang2024game}.

\textit{2) 4WD-Vehicle Model}: The state vector of the 4WD vehicle model at time step $t$ is $x^{(t)}=[p_x, p_y, \theta]$ and its control input vector is $u^{(t)}=[v_L, v_R]$ which represents the linear velocity of the left and right wheels. Let the distance between the left and right wheels be $L$, the linear and angular velocity of the vehicle are $v = \frac{1}{2} (v_L + v_R)$ and $\omega = \frac{1}{L} (v_R - v_L)$. Then, the vehicle's state is updated by $\dot{x}^{(t)}=A x^{(t)} + B  u^{(t)}$ where: 
\[
A = \begin{bmatrix}
0 & 0 & -v\sin\theta \\
0 & 0 & v\cos\theta \\
0 & 0 & 0
\end{bmatrix},
B = \begin{bmatrix}
\frac{\cos\theta}{2} & \frac{\cos\theta}{2} \\
\frac{\sin\theta}{2} & \frac{\sin\theta}{2} \\
-\frac{1}{L}         & \frac{1}{L}
\end{bmatrix}.
\] 

\section*{Appendix -- Hyperparameter Selection}

All hyperparameters in our framework are selected by a systematic grid search aimed at optimizing the task metrics: {zero collision} (hard requirement), {minimum tracking cost}, and {maximum efficiency} (shorter arrival time).
We predefine discrete grids for each hyperparameter, roll out each configuration over multiple random seeds, and evaluate:
\textit{(i)} collision ratio (must be $0$), \textit{(ii)} tracking cost (lower is better), and \textit{(iii)} efficiency, indicated by distance to goal at $t=T-5$ (lower is better).
Selection follows a \emph{safety-first} lexicographic rule:
we first discard all configurations that exhibit any collision. Among the remaining collision-free configurations, we select the one with the smallest tracking cost $J$, and if multiple remain, we break ties by choosing the configuration with higher efficiency.
The final selected values for each scenario are listed in the manuscript.

\begin{figure}[h]
\centering
\includegraphics[width=\linewidth]{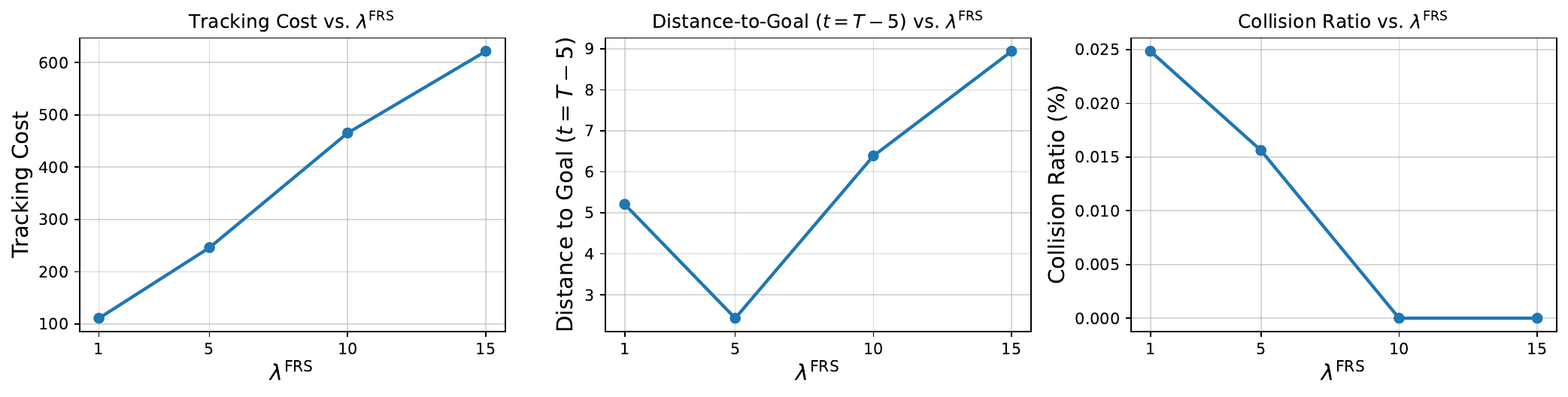}
\caption{Tuning results for the MA-FRS penalty weight $\lambda^{\mathrm{FRS}}$. (Left) Tracking Cost, (Middle) Distance-to-Goal (Efficiency), and (Right) Collision Ratio.}
\label{fig:metrics_vs_lambda_frs}
\end{figure}

\textbf{Demonstrative Example: $\lambda^\text{FRS}$. }
To make the process concrete, we select $\lambda^\text{FRS}$ as a demonstrative example. We fix $(Q_i,R_i,\lambda^{\mathrm V})$ at their reported values and sweep
$\lambda^{\mathrm{FRS}}$ on a grid $\{1, 5, 10, 15\}$.
For each $\lambda^{\mathrm{FRS}}$ value, we run $50$ rollouts and report the collision ratio, tracking cost, and efficiency. The results are shown in Fig.~\ref{fig:metrics_vs_lambda_frs}.

The tuning process reveals a typical outcome: When $\lambda^{\mathrm{FRS}}$ is too small (e.g., 1 or 5), the marginal reduction in tracking cost can outweigh the safety penalty. The optimizer then tolerates MA-FRS overlaps, leading to a nonzero collision ratio. As $\lambda^{\mathrm{FRS}}$ increases (e.g., 10 or 15), the gradient of the safety penalty dominates near potential overlaps at which point the observed collision ratio becomes $0\%$. Pushing $\lambda^{\mathrm{FRS}}$ further (e.g., 15) prioritizes clearance and introduces conservatism, which slightly increases tracking cost and reduces efficiency (longer paths). In our sweep, we therefore choose $\lambda^{\mathrm{FRS}}=10$ as the smallest value that achieves $0\%$ collision while maintaining lower tracking cost and good efficiency within the collision-free regime.

\section*{Appendix -- Scalability Analysis}
To analyze the impact of the local cost approximation on computational scalability, this appendix quantifies the effect of the localized approximation used in the ND-iBR scheme. We compare the neighborhood size of our MA-FRS–based localized ND-iBR ({neighbors only}) against a theoretical {all-to-all coupling} variant (i.e., $|\tilde{\mathcal{N}}_i|=N{-}1$ for all $i$). We quantify the {time-and-agent-averaged} neighborhood size over the planning horizon as:
\begin{equation}
\overline{|\tilde{\mathcal N}|}
:= \frac{1}{NT}\sum_{t=0}^{T-1}\sum_{i=1}^{N} \big|\tilde{\mathcal N}_i^{(t)}\big|, 
\end{equation}
while for the all-to-all baseline, $\overline{|\tilde{\mathcal N}|}=N{-}1$.

\begin{table}[ht]
\centering
\caption{Average number of neighbors per agent.}
\label{tab:avg_neighbors}
\begin{tabular}{lcccccc}
\toprule
\textbf{Agents $N$} & 3 & 4 & 5 & 6 & 7 & 8 \\
\midrule
\textbf{Avg.\ neighbors} & 0.67 & 1.43 & 1.76 & 1.80 & 2.70 & 2.76 \\
\textbf{$\frac{N-1}{\text{Avg.}}$} & 2.98 & 2.09 & 2.27 & 2.78 & 2.22 & 2.54 \\
\bottomrule
\end{tabular}

\bigskip

\begin{tabular}{lccccccc}
\toprule
\textbf{Agents $N$} & 9 & 10 & 11 & 12 & 13 & 14 & 15 \\
\midrule
\textbf{Avg.\ neighbors} & 3.28 & 3.57 & 3.50 & 4.51 & 5.23 & 5.24 & 5.51 \\
\textbf{$\frac{N-1}{\text{Avg.}}$} & 2.44 & 2.52 & 2.85 & 2.44 & 2.29 & 2.48 & 2.72 \\
\bottomrule
\end{tabular}
\end{table}

As shown in Table~\ref{tab:avg_neighbors}, $\overline{|\tilde{\mathcal N}|}$ grows sublinearly with $N$ (from $0.67$ at $N{=}3$ to $5.51$ at $N{=}15$), remaining far below $(N{-}1)$ at all sizes . Consequently, the effective coupling dimension per agent per step is reduced by a factor $(N{-}1)/\overline{|\tilde{\mathcal N}|}$ ranging from $2.09$ to $2.98$ across $N{=}3$ to $15$. Operationally, this reduction means that each planning step instantiates and differentiates fewer {pairwise soft-penalty} terms (MA-FRS overlap penalties). Since an agent only penalizes its current neighbors, the total number of soft constraints scales as $O\!\big(N\,\overline{|\tilde{\mathcal N}|}\big)$ rather than $O\!\big(N(N{-}1)\big)$. Fewer coupling terms lead to smaller KKT systems and faster solutions.

% \vspace{-0.5em}

\bibliographystyle{IEEEtran}
\bibliography{reference_abbreviated}

\end{document}